\pdfoutput=1

\documentclass[11pt]{article}

\usepackage[preprint]{acl}

\usepackage{times}
\usepackage{latexsym}

\usepackage{subcaption}
\usepackage{listings}
\usepackage{multirow}
\usepackage{enumitem}
\usepackage{amsmath}
\usepackage{graphicx,array,booktabs,url,amssymb}
\usepackage{xspace}

\newcommand{\dataset}{\textsc{ProAssist}\xspace}

\usepackage[T1]{fontenc}

\usepackage[utf8]{inputenc}

\usepackage{microtype}

\usepackage{inconsolata}

\usepackage{graphicx}

\title{Proactive Assistant Dialogue Generation from Streaming Egocentric Videos}

\author{
 \textbf{Yichi Zhang\textsuperscript{1,2,}\thanks{~Work partially done at Meta.}},
 \textbf{Xin Luna Dong\textsuperscript{1}},
 \textbf{Zhaojiang Lin\textsuperscript{1}},
 \textbf{Andrea Madotto\textsuperscript{1}},
\\
 \textbf{Anuj Kumar\textsuperscript{1}},
 \textbf{Babak Damavandi\textsuperscript{1}},
 \textbf{Joyce Chai\textsuperscript{2}},
 \textbf{Seungwhan Moon\textsuperscript{1}}
\\
\\
 \textsuperscript{1}Meta~~~
 \textsuperscript{2}University of Michigan
\\
 \small{
   {Correspondence:} \href{mailto:email@domain}{zhangyic@umich.edu}
 }
}

\begin{document}
\maketitle

\begin{abstract}
Recent advances in conversational AI have been substantial, but developing real-time systems for perceptual task guidance remains challenging. These systems must provide interactive, proactive assistance based on streaming visual inputs, yet their development is constrained by the costly and labor-intensive process of data collection and system evaluation. To address these limitations, we present a comprehensive framework with three key contributions. First, we introduce a novel data curation pipeline that synthesizes dialogues from annotated egocentric videos, resulting in \dataset, a large-scale synthetic dialogue dataset spanning multiple domains. Second, we develop a suite of automatic evaluation metrics, validated through extensive human studies. Third, we propose an end-to-end model that processes streaming video inputs to generate contextually appropriate responses, incorporating novel techniques for handling data imbalance and long-duration videos. This work lays the foundation for developing real-time, proactive AI assistants capable of guiding users through diverse tasks. 
Project page: \url{https://pro-assist.github.io/}


\end{abstract}

\vspace{-10pt}
\section{Introduction}
\vspace{-5pt}
\label{sec:intro}

\begin{figure}[t]
    \centering
    \includegraphics[width=0.88\linewidth]{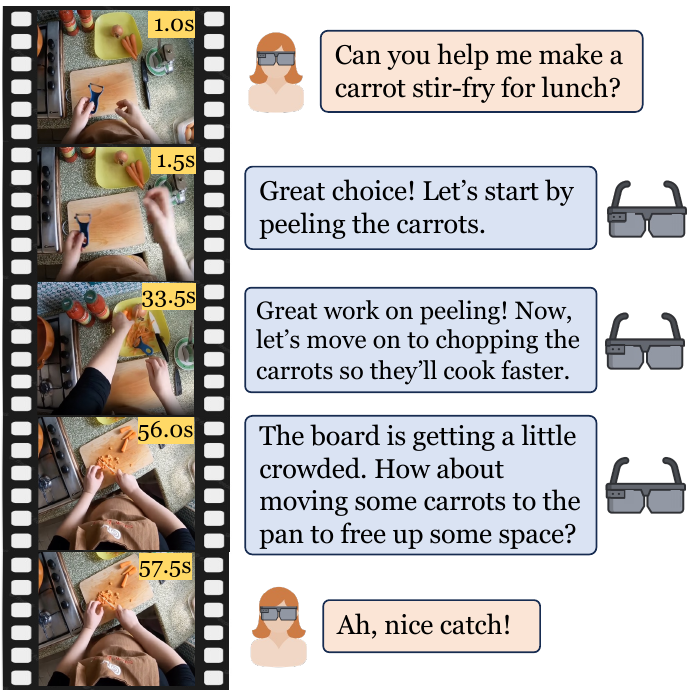}
    \vspace{-10pt}
    \caption{An example conversation between user and their task assistant. The assistant receives real-time video streams from the user's perspective, and provides proactive guidance to assist with the task. Images excerpted from Ego4D \cite{grauman2022ego4d}.\vspace{-20pt}}
    \label{fig:teaser}
\end{figure}

Recent advances in multimodal language models have transformed various aspects of human-AI interaction \cite{achiam2023gpt,team2024gemini,dubey2024llama}. However, developing AI systems capable of providing real-time, interactive guidance for physical tasks remains a significant challenge \cite{bao2023can}. Ideally, as illustrated in Figure \ref{fig:teaser}, such an assistant should proactively guide users through each step of a task based on a high-level goal, determining both when and how to communicate through continuous processing of the environment and understanding of task objectives. This requires the system to handle streaming video inputs while simultaneously managing diverse user interactions including requests, questions, and comments, and offering timely guidance upon detecting the completion of each task step. The dual challenge of determining both appropriate response timing and content through real-time processing of long-horizon video inputs makes task guidance a particularly complex problem.

Despite significant progress in component technology such as 
action recognition \cite{damen2020epic,grauman2022ego4d}, mistake detection \cite{sener2022assembly101,wang2023holoassist,lee2024error}, and question answering \cite{wong2022assistq,ilaslan2023gazevqa}, 
we are still far from enabling the holistic ability to generate appropriate dialogue responses for task guidance. Two major challenges have hindered progress toward this goal: the lack of large-scale and diverse training data, as existing datasets are primarily constrained by labor-intensive Wizard-of-Oz setups limited to single domains \cite{bao2023can,wang2023holoassist}; and the lack of scalable evaluation frameworks for assistant dialogue generation that could serve as efficient proxies for human evaluation to support rapid and reproducible model comparisons during system development.

To address these limitations, we propose a new problem of proactive assistant dialogue generation from streaming videos and develop comprehensive resources to approach this problem.
We introduce an automated approach for synthesizing task-oriented dialogues from well-annotated egocentric video datasets \cite{song2024ego4d,huang2024egoexolearn,grauman2024ego,damen2020epic,sener2022assembly101}, resulting in \dataset -- a large-scale synthetic dialogue dataset containing 30,135 dialogues across 479  hours of video in cooking, object manipulation, assembly, and laboratory domains. Our method leverages state-of-the-art large language models \cite{achiam2023gpt,TheC3,dubey2024llama} to generate realistic assistant-user interactions, using detailed timestamped video descriptions to maintain temporal alignment. For systematic evaluation, we propose two complementary automatic metrics: a pairwise approach based on sentence matching and an end-to-end approach utilizing LLM-as-a-Judge \cite{zheng2023judging}. Through extensive human studies, we validate both the quality of our synthetic data and the alignment between our proposed metrics and human judgment, establishing \dataset as a reliable benchmarking resource.

Based on \dataset, we develop an end-to-end multimodal large language model (MLLM) for generating contextually appropriate responses from streaming video inputs. Building upon the VideoLLM-Online architecture \cite{chen2024videollm}, we introduce two key innovations: negative frame sub-sampling to improve response timing decisions, and iterative progress summarization to enable efficient processing of long video sequences. Our experimental results demonstrate the effectiveness of these modeling techniques while providing valuable insights into the complexities of perceptual task guidance.

\vspace{-3pt}
\section{Related Work}
\label{sec:related_work}

\vspace{-3pt}
\paragraph{Interactive Assistant for Task Guidance.}
\vspace{-3pt}
Task guidance systems have evolved from early rule-based policies \cite{ockerman1998preliminary,ockerman2000review} to perception-enabled but task-specific solutions \cite{leelasawassuk2017automated,reyes2020mixed,lu2019higs,wang2016multi,sato2014mimicook,bao2023can}. Recent research has primarily focused on developing components for single-domain systems \cite{wang2023holoassist}, including environment understanding \cite{wong2022assistq,ilaslan2023gazevqa}, user behavior analysis \cite{damen2020epic,grauman2022ego4d,huang2024egoexolearn}, and mistake detection \cite{sener2022assembly101,lee2024error,peddi2023captaincook4d}. Our work differs by evaluating end-to-end dialogue generation capabilities of a general-purpose system across multiple domains.

\vspace{-3pt}
\paragraph{Synthetic Dialogue Generation.}
\vspace{-3pt}
Synthetic dialogue generation has proven effective for creating large-scale datasets in both text-only \cite{shah2018building,mohapatra2021simulated,rastogi2020towards} and multimodal scenarios \cite{kottur2021simmc,wuetal2023simmc,zhan2024going,moon2020situated}. Recent LLMs have shown remarkable capabilities in simulating human behavior \cite{park2023generative}, particularly valuable for low-resource scenarios \cite{li2022controllable,abdullin2024synthetic,chen2023places}. While previous work has demonstrated the feasibility of generating dialogues about visual content using structured descriptions \cite{liu2024visual,maaz2023video,luo2023valley,chen2024videollm}, our approach uniquely employs a dedicated LLM pipeline to generate natural assistant-user interactions for egocentric task completion videos.

\vspace{-5pt}
\paragraph{Multimodal Dialogue Modeling.}
\vspace{-3pt}
The success of large language models (LLMs) \cite{brown2020language,ouyang2022training} has led to the development of multimodal variants (MLLMs) capable of handling both image-based \cite{alayrac2022flamingo,liu2024visual,li2023blip,zhu2023minigpt} and video-based dialogues \cite{li2023videochat,lin2023video,maaz2023video,song2024moviechat,yang2023vid2seq,zhang2023video}. However, these models typically operate in an offline setting with access to complete videos.
While VideoLLM-Online \cite{chen2024videollm} represents a significant step toward online video processing, it primarily focuses on short video clips. Our work extends the model with novel techniques specifically designed for task guidance scenarios, introducing mechanisms for response timing decisions and efficient processing of long-horizon videos.
\vspace{-5pt}
\section{Proactive Assistant Dialogue Generation from Streaming Videos}
\label{sec:proact_dataset}
\vspace{-5pt}


\begin{figure*}[t]
    \centering
    \includegraphics[width=0.99\linewidth]{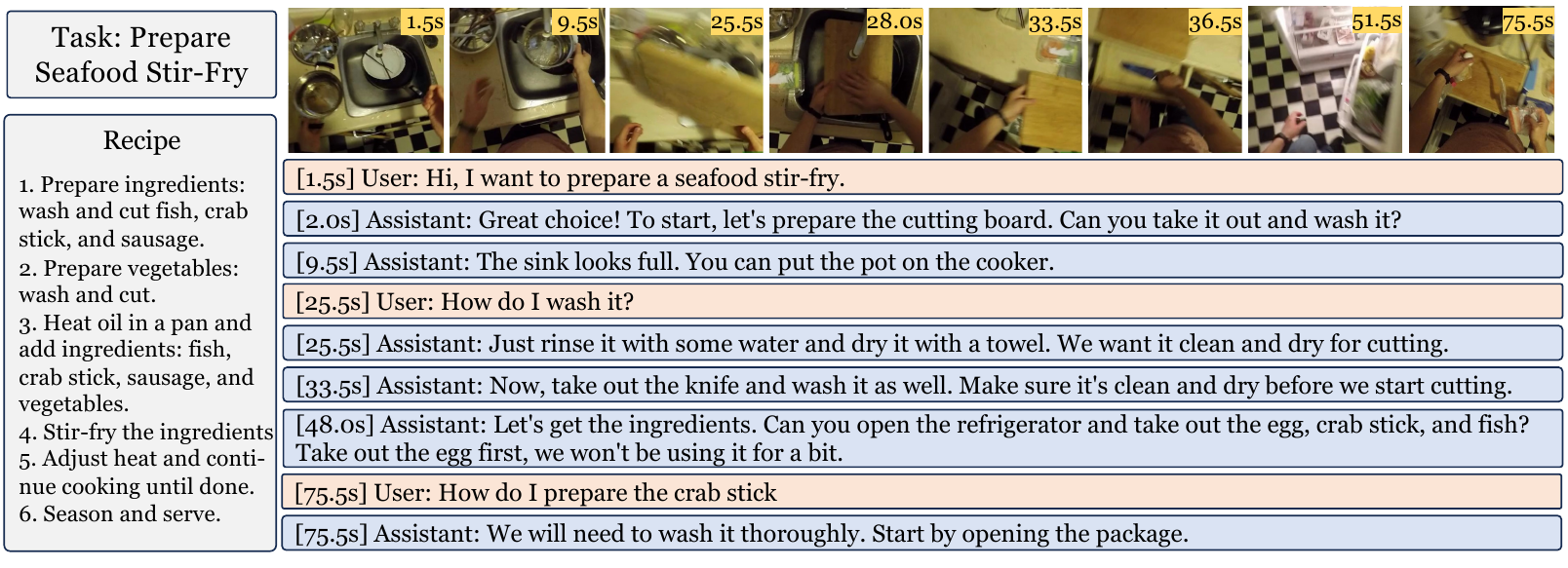}
    \vspace{-10pt}
    \caption{A cooking task example from \dataset, derived from EpicKitchen. The task goal, recpie and dialogue are generated through our synthetic data curation pipeline (dialogue partially shown due to space constraints). \vspace{-15pt}}
    \label{fig:data-sample}
\end{figure*}


\subsection{Problem Definition}
\label{sec:stream_v2t_gen}

The goal of a proactive assistant system is to generate prompt, appropriate, and helpful guidance from egocentric video streams in real time. We formulate this problem as a streaming video-to-dialogue generation task. Given a video stream of $T$ frames, our objective is to generate a sequence of assistant responses $s_{1:T}$ that maximizes the conditional probability:
\vspace{-8pt}
\begin{equation}
\label{eq1}
s_{1:T} = \arg\max \prod_tP(s_t|v_{1:t}, s_{1:t-1}, k)
\vspace{-12pt}
\end{equation}
where $s_t$ represents the assistant's response at time step $t$ (either a textual message to user or $\varnothing$ for keeping silence), $v_t$ denotes the multimodal input including the video frame and optional user utterance input, and $k$ represents optional task knowledge (e.g., a recipe). The task begins when the user provides a goal through text input. When $k$ is provided, we term this knowledge-conditioned evaluation, reflecting a realistic retrieval-augmented setup for real-world systems. This formulation requires the assistant to determine both when to speak and what to say based on current visual context, dialogue history, and task understanding.


\vspace{-5pt}
\subsection{\hspace{-1pt}\dataset: A Synthetic Dialogue Dataset}
\vspace{-5pt}



We now describe our approach for the creation of \dataset. We first collect egocentric videos that are extensively annotated with timestamped user action descriptions from six public dataset: Ego4D-Goalstep \cite{grauman2022ego4d,song2024ego4d}, EpicKitchen \cite{damen2020epic}, HoloAssist \cite{wang2023holoassist}, Assembly101 \cite{sener2022assembly101}, EgoExoLearn \cite{huang2024egoexolearn}, and WTaG \cite{bao2023can}. The annotations are processed into a standardized format, \texttt{[t]<description>}, where \texttt{[t]} represents a timestamp or time span.
Additional annotations such as high-level task step and error correction labels are similarly formatted and inserted in chronological order whenever available.
This unified representation enables LLMs to effectively understand ongoing activities at each time step.

Building on these annotations, we design a data curation pipeline consisting of the following steps:

\noindent\textbf{1. Task Goal and Recipe Generation}: We first prompt the LLM to summarize the task goal and generate a task recipe based on the video descriptions. This step will be skipped if the dataset already includes these elements (e.g., WTaG). The generated goal serves as the initial user input describing the task, while the recipe supports knowledge-conditioned evaluation (\S\ref{sec:kce}).
     
\noindent\textbf{2. Video Pre-Filtering}: Non-procedural, multi-tasking, or incompletely annotated videos are filtered out to ensure dataset quality.

\noindent\textbf{3. Multi-Round Dialogue Generation}: Dialogues are generated using three types of user behavior: {\em no talk} (i.e., silent except for giving the goal), {\em talk some}  (occasional task-related questions), and {\em talk more} (frequent conversational interactions). Inputs include the goal, video descriptions, and user behavior type. To handle long video descriptions, we adopt a multi-round generation approach, dividing videos into chunks and generating dialogues incrementally to stay within the LLM’s context window. Afterward, we prompt the LLM for a refinement pass to improve dialogue naturalness and coherence. At this step, we generate 10 dialogues per video, distributed across user types in a 2:4:4 ratio.

\noindent\textbf{4. Dialogue Annotation}: The generated dialogues are then labeled by LLM, including assistant intent (instruction, mistake correction, feedback) and response type (responsive or proactive). Additionally, we also generate a summary of progress at each assistant turn for the user's progress so far, which will be used to support the iterative progress summarization approach (\S\ref{sec:summarization}).

\noindent\textbf{5. Automatic Quality Evaluation and Post-Filtering}: We perform automatic evaluations to ensure the dialogues meet high standards, assessing timing precision, task step coverage, and assistant responsiveness. Low-quality dialogues are filtered out from the training set. For the validation split, we retain only the highest-scoring dialogue per user type, splitting them evenly into validation and test sets. This process removes approximately 25\% of dialogues and 41 hours of video. Final data statistics are shown in Table \ref{tab:statistics}.

We leverage LLaMA-3.1-70B-Instruct \cite{dubey2024llama} as the LLM for all the aforementioned steps. An example dialogue generated through this pipeline is shown in Figure \ref{fig:data-sample}. 
To ensure the safety of our generated dataset, we applied the LLaMA-Guard-3-8B model\footnote{https://huggingface.co/meta-llama/Llama-Guard-3-8B} to all generated dialogues for safety classfication. The classifier flagged 17 instances (0.05\%) as potentially unsafe. Upon manual inspection, we found no actual issues in these flagged cases, indicating that they were likely false positives. 
More details including the prompt for each step, data distributions and dialogue examples are available in the Appendix.



\begin{table}[h!]
\vspace{-5pt}
\centering
\resizebox{\columnwidth}{!}{
\addtolength{\tabcolsep}{-0.3em}
\begin{tabular}{lcccc}
\toprule
\textbf{Subset} & Video Hour & \#Videos & \#Dialogues \\
\midrule
Ego4D        & 136.6 / 11.6 / 13.2  & 382 / 32 / 33  & 3182 / 96 / 99 \\
HoloAssist   & 107.0 / 7.6 / 6.8   & 1436 / 97 / 97 & 7052 / 291 / 291 \\
EgoExoLearn  & 68.5 / 8.8 / 9.4    & 321 / 41 / 41  & 3210 / 123 / 123 \\
Assembly101  & 43.1 / 6.9 /7.2   & 756 / 112 / 112 & 7492 / 336 / 336 \\
EpicKitchens & 34.0 / 4.2 / 4.1    & 320 / 50 / 50  & 6376 / 150 / 150 \\
WTaG         & 7.1 / 1.2 / 1.3     & 40 / 7 / 7     & 786 / 21 / 21 \\
\midrule
Total & 478.7 & 3934 & 30135 \\
\bottomrule
\end{tabular}}
\vspace{-5pt}
\caption{Data statistics of \dataset for train/validation/test splits. More statistics in Appendix \ref{supp:statistics}. \vspace{-10pt}}
\label{tab:statistics}
\end{table}



\vspace{-5pt}
\section{Evaluation of Proactive Task Assistant}
\vspace{-5pt}

\begin{figure*}[t]
    \centering
    \includegraphics[width=0.98\linewidth]{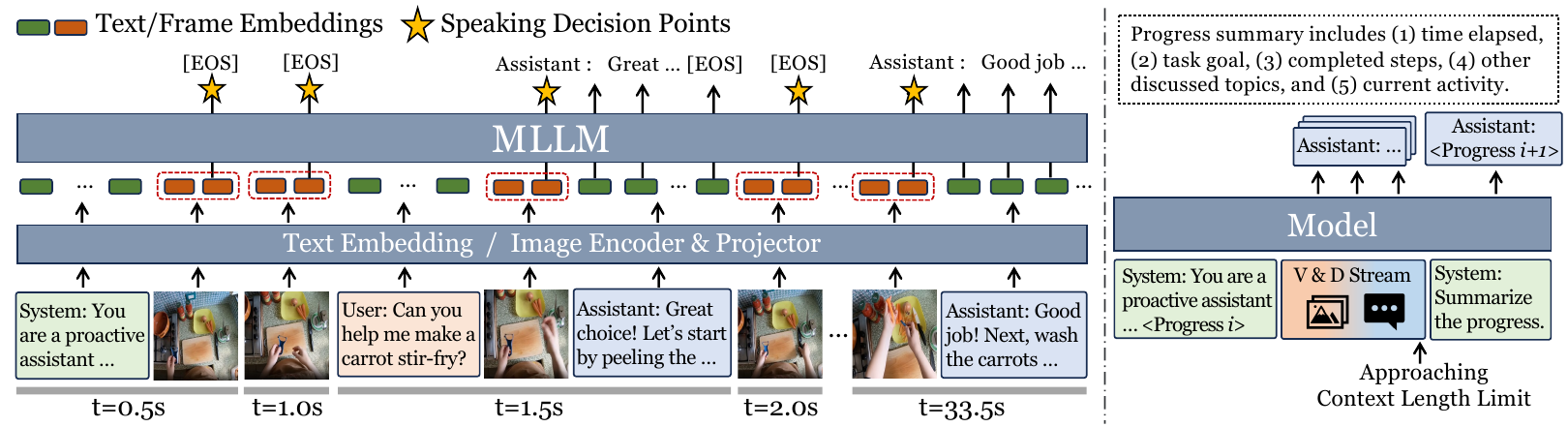}
    \vspace{-5pt}
    \caption{\textbf{Left}: Streaming video-to-dialogue generation using VideoLLM-Online. The model processes live video frames and optional textual inputs to decide whether to speak or remain silent at designated decision points (yellow stars), and autogressively generates assistant responses as needed. Learning when to speak faces significant class imbalance due to the sparsity of speaking frames. \textbf{Right}: Illustration of iterative progress summarization. When approaching its context length limit, the model generates a concise task progress summary, then restarts generation with this summary incorporated into a new system prompt.\vspace{-15pt}}
    \label{fig:model}
\end{figure*}



\label{sec:evaluation}

Evaluating interactive dialogue systems is inherently challenging \cite{deriu2021survey}, particularly for proactive task guidance where both response timing and content must be assessed. While direct human evaluation through system interaction would be ideal, the high cost makes it impractical for large-scale benchmarking, especially during rapid development cycles. We therefore propose an offline evaluation framework that enables efficient, automatic dataset-based assessment.

Our framework aims to measure the overall helpfulness of a system's sequential predictions $s_{1:T}$ for a video stream $v_{1:T}$ (as defined in Eq.\ref{eq1}) by comparing them against ground-truth dialogue $\hat{s}_{1:T}$. This comparison is challenging due to the potential long task horizon and the need to evaluate predictions that may differ from ground-truth in both timing and content. Below, we introduce two evaluation metrics designed to address these challenges.

\vspace{-3pt}
\paragraph{Pairwise Evaluation via Sentence Matching}
\vspace{-5pt}
Our first metric evaluates system performance by matching each predicted utterance with semantically similar and temporally aligned reference utterances. Based on these matches, we compute three metrics: \textit{precision} (matched predictions over total predictions), \textit{recall} (matched predictions over total references), and \textit{F1} (their harmonic mean). To identify optimal matches, we apply bipartite matching based on a cost matrix combining both semantic and temporal alignment costs. The semantic cost between predictions $s_i$ and references $\hat{s}_j$ is defined as:
\vspace{-8pt}
\begin{equation*}
\vspace{-3pt}
    s(i,j)=
\begin{cases}
1 &\text{if } \hat{s}_j = \varnothing \\
1-sim(e(s_i), e(\hat{s}_j)) & \text{else}
\end{cases}
\label{eq2}
\vspace{-3pt}
\end{equation*}
where $e(\cdot)$ is the sentence embedding function and $sim(\cdot)$ denotes cosine similarity. The temporal cost encourages matching with temporally proximate messages:
\vspace{-8pt}
\begin{equation*}
\vspace{-3pt}
    d(i,j)=
\begin{cases}
|i-j|^p &\text{if } i-j\in[-L,R] \\
\infty & \text{else}
\end{cases}
\label{eq3}
\vspace{-3pt}
\end{equation*}
where $p$ controls the cost increase rate with time difference, while $R$ and $L$ define maximum allowable time differences for predictions preceding or following references. We set $R < L$ to favor earlier predictions, preventing the model from exploiting future frame information. The final matches are computed using the LAPJVsp algorithm \cite{jonker1988shortest} with a weighted sum of both costs. More details are provided in the Appendix.

\vspace{-3pt}
\paragraph{End-to-End Evaluation via LLM-as-a-Judge}
\vspace{-3pt}
Our ultimate goal is to evaluate the assistant's overall usefulness to the user. While the pairwise matching approach approximates this through prediction-reference similarity, it cannot capture the flexibility of different guidance strategies. Drawing inspiration from recent LLM-based evaluation approaches \cite{zheng2023judging,liu2023g,maaz2023video}, we propose using LLMs to directly assess the quality of the overall assistance experience. Given timestamped predictions and reference dialogues, we prompt an LLM to evaluate system performance across four dimensions: \textit{correctness} of guidance and feedback, \textit{appropriateness} of response timing, \textit{efficiency} of information delivery, and \textit{overall} helpfulness. Each aspect is rated on a 5-point Likert scale from ``very poor'' to ``excellent''. For reliability, we average scores from three independent runs. The complete evaluation prompt is provided in the Appendix.

Note that our metrics are generally applicable beyond \dataset to any dataset with ground-truth dialogues. Moreover, the pairwise evaluation can be applied to other streaming video-to-text tasks requiring joint assessment of timing and content, such as online action narration (see \S\ref{sec:narration_task}).
\vspace{-5pt}
\section{Proactive Assistant Dialogue Modeling}
\label{sec:modeling}
\vspace{-5pt}

Next, we present our exploration into developing a functional proactive assistant dialogue generation model. We begin with an analysis of existing models to assess their feasibility in addressing our problem. Then, we describe how we enhance a baseline model with two novel techniques to enable it to tackle the unique challenges involved.

\subsection{Feasibility Analysis of Existing Models}
Streaming video-to-dialogue generation poses unique modeling challenges to real-time video processing and online text generation. Most existing MLLMs \cite{lin2023video,maaz2023video,zhang2023video,li2025llama,moon2024anymal,he2024ma,weng2025longvlm,wang2024gpt4video} are designed for offline scenarios where the complete video is available beforehand, making them unsuitable for our setup. While state-of-the-art proprietary MLLMs \cite{achiam2023gpt,TheC3,team2024gemini} can process interleaved image-text inputs, they suffer from high API latency and cost\footnote{For example, deciding when and what to say at 2 FPS for a 30-minute video requires making 6000 API calls.} and often struggle to determine appropriate response timing \cite{chen2024videollm}.

VideoLLM-Online \cite{chen2024videollm} offers a viable baseline for our task, as it specifically handles live-streamed video inputs. As shown in Figure \ref{fig:model}, the model processes interleaved video frames and textual inputs by encoding frames into visual tokens through a frozen pretrained image encoder and a tunable projector layer. For each frame, it predicts whether to respond at the last visual token position, generating \texttt{[EOS]} to remain silent or initiating response generation otherwise. To enable real-time interaction, it employs a compact frame representation of 1-10 tokens, significantly fewer than mainstream MLLMs.

However, VideoLLM-Online faces two key limitations in our task guidance scenario: the difficulty of learning when to speak due to the sparsity of speaking moments, and the inability to handle long-horizon tasks due to context window constraints. In the following sections, we present our enhanced model that addresses these challenges through two novel techniques.

\vspace{-5pt}
\subsection{\hspace*{-1pt}Learning When to Speak under Imbalance}
\label{sec:subsampling}
\vspace{-5pt}

Learning when to speak can be framed as a sequence of binary decisions, where at each step the model must choose between speaking and remaining silent. We denote frames requiring responses as positive samples and those requiring silence as negative samples. As shown in Figure \ref{fig:model} (left), the speaking decision points (yellow stars) demonstrate a significant imbalance, with far more negative samples (predicting \texttt{[EOS]}) than positive ones (predicting \texttt{Assistant}). This imbalance creates a challenging learning problem, as directly optimizing cross-entropy on the original distribution leads to a classifier biased toward silence.

We propose Negative Frame Sub-sampling (NFS) to address this challenge. During training, we compute gradients only for positive frames and a uniformly sampled subset of negative frames, comprising a proportion $\rho$ of total negative samples. The loss remains unchanged for non-decision positions to maintain response generation capability. This approach can be efficiently implemented by adjusting the gradient computation mask without modifying model inputs. Furthermore, dynamically resampling negative samples each epoch ensures all positions can potentially contribute to learning over time, enhancing model robustness.


\vspace{-5pt}
\subsection{Iterative Progress Summarization}
\label{sec:summarization}
\vspace{-5pt}


Long-horizon tasks (e.g., hour-long videos) challenge models in tracking goals and progress over time during both training and inference. Hardware constraints (e.g., GPU memory) during training enforce fixed-length sequence processing (L), forcing truncation of longer samples\footnote{For example, 86\% of samples from the Ego4D subset of \dataset must be truncated when L=4096 and each frame is encoded as 10 tokens.}, causing substantial information loss and hindering the learning of long-horizon task progressions. During inference, context length limitations similarly restrict processing to tasks within the model's training window.


We introduce Iterative Progress Summarization (IPS) to overcome these issues, enabling continuous task tracking via dynamic memory compression. As shown in Figure \ref{fig:model} (right), when approaching context limits, the model generates a concise, task-relevant progress summary. Generation then resumes with this summary incorporated into the initial system prompt for the next processing segment. In training, long videos are preprocessed into context-fitting chunks with summaries carried forward. Critically, unlike methods requiring specialized training \cite{wang2023recursively, chevalier2023adapting}, IPS integrates with standard LLM training, enabling our model to handle potentially infinite-length video streams while maintaining task and progress tracking.
\vspace{-5pt}
\section{Experiments}
\label{sec:experiment}

\vspace{-5pt}
\subsection{Experiment Setups}
\vspace{-3pt}
\label{sec:narration_task}
\noindent\textbf{Baseline Task.} While we primarily evaluate our model on proactive assistant dialogue generation, we also include egocentric action narration as a baseline task, where the model describes the camera wearer's actions in real-time. Action narration serves as a simpler variant of streaming video-to-text generation that mainly requires visual perception capabilities. By comparing performance between action narration and dialogue generation, we can better understand how well our model handles capabilities beyond visual perception, such as situational reasoning and progress tracking, which are essential for effective task guidance.


\begin{figure}[t]
    \centering
    \includegraphics[width=0.98\linewidth]{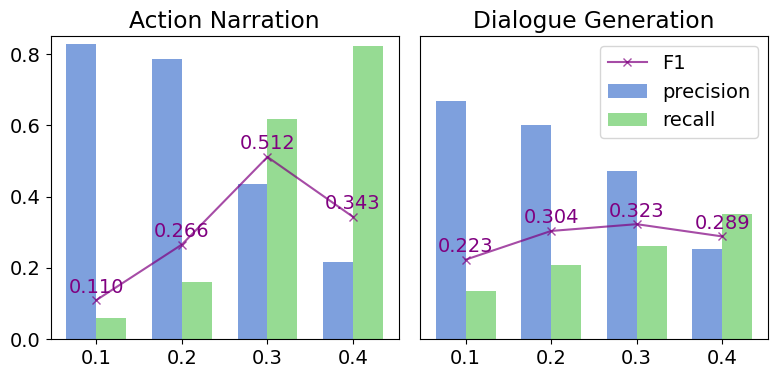}
    \vspace{-10pt}
    \caption{Model performance under different speaking decision threshold. The trade-off between precision and recall exists across both tasks.\vspace{-15pt}}
    \label{fig:pr-tradeoff}
\end{figure}

\noindent\textbf{Knowledge-Conditioned Evaluation.}
\label{sec:kce}
We introduce a knowledge-conditioned evaluation setup where the model receives task-specific instructions (e.g., recipes) as a system prompt after the user states their goal. This setup mirrors real-world scenarios where assistants access user-provided recipes or retrieved knowledge to offer guidance.

\noindent\textbf{Speaking Decision Threshold.}
\label{sec:speaking_decision}
At inference time, we convert the model's probabilistic token predictions into binary decisions using a threshold $\theta$: the model remains silent if the probability of \texttt{[EOS]} exceeds $\theta$. Our experiments show that model performance is highly sensitive to $\theta$, with a clear precision-recall tradeoff (Figure \ref{fig:pr-tradeoff}). We use the $\theta$ with the highest validation F1 score for testing.

\noindent\textbf{Model Variants.}
We implement three variants of VideoLLM-Online, with different number of visual tokens per frame: $I=1, 5, 10$. The model is intuitively better at visual perception with more tokens, with a cost of computationally more expensive. 

\noindent\textbf{Implementation Details.}
We use LLaMA-3.1-8B-Instruct~\cite{dubey2024llama} as the backbone and SigLIP-SO400M \cite{zhai2023sigmoid} as the frame encoder for VideoLLM-Online. For training, we adopt a single stage training 
on mixed data of dialogues both with and without knowledge from \dataset, Ego4D online action narration, and some auxiliary vision-language datasets, resulting in a single model that can be tested with different setup and tasks. See the Appendix for more details.

\begin{table}[t!]
    \centering
    \resizebox{\columnwidth}{!}{
    \addtolength{\tabcolsep}{-0.3em}
    \begin{tabular}{lcccc}
        \toprule
         & Correctness & Helpfulness & Alignment & Naturalness \\
        \midrule
        All & 3.27 {\scriptsize$\pm$ 0.79} & 3.46 {\scriptsize$\pm$ 0.77} & 2.91 {\scriptsize$\pm$ 1.00} & 3.54 {\scriptsize$\pm$ 0.70} \\
        - No Talk & 3.27 {\scriptsize$\pm$ 0.70} & 3.47 {\scriptsize$\pm$ 0.74} & 2.75 {\scriptsize$\pm$ 0.96} & 3.32 {\scriptsize$\pm$ 0.79} \\
        - Talk Some & 3.23 {\scriptsize$\pm$ 0.86} & 3.37 {\scriptsize$\pm$ 0.80} & 2.93 {\scriptsize$\pm$ 1.01} & 3.50 {\scriptsize$\pm$ 0.74} \\
        - Talk More & 3.32 {\scriptsize$\pm$ 0.79} & 3.53 {\scriptsize$\pm$ 0.76} & 3.05 {\scriptsize$\pm$ 0.99} & 3.80 {\scriptsize$\pm$ 0.44} \\
        \midrule
        HoloAssist-Gen & 3.15 {\scriptsize$\pm$ 0.91} & 3.40 {\scriptsize$\pm$ 0.49} & 2.65 {\scriptsize$\pm$ 0.91} & 3.60 {\scriptsize$\pm$ 0.73} \\
        HoloAssist-Human & 2.88 {\scriptsize$\pm$ 1.05} & 2.62 {\scriptsize$\pm$ 1.11} & 2.75 {\scriptsize$\pm$ 1.09} & 2.50 {\scriptsize$\pm$ 1.32} \\
        \hline
        WTaG-Gen & 3.50 {\scriptsize$\pm$ 0.59} & 3.50 {\scriptsize$\pm$ 0.92} & 3.15 {\scriptsize$\pm$ 1.11} & 3.65 {\scriptsize$\pm$ 0.73} \\
        WTaG-Human & 3.60 {\scriptsize$\pm$ 0.49} & 3.60 {\scriptsize$\pm$ 0.66} & 3.60 {\scriptsize$\pm$ 0.66} & 3.60 {\scriptsize$\pm$ 0.66} \\
        \bottomrule
    \end{tabular}}
    \vspace{-10pt}
    \caption{Human evaluation of the generated dialogue quality. For HoloAssist and WTaG where human-collected dialogues are available, we evaluate them using the same approach for a side-by-side comparison with our generated dialogues.\vspace{-12pt}}
    \label{table:human_eval_dialog}
\end{table}

\vspace{-5pt}
\subsection{Dialogue Quality of \dataset}
\vspace{-5pt}

To validate \dataset as a reliable resource for studying proactive assistant dialogue generation, we conducted a comprehensive human evaluation of the synthetic dialogues. We uniformly sampled 100 dialogues from the test split across all six data subsets, covering three user types (i.e., {\em no talk}, {\em talk some}, and {\em talk more}). Two annotators evaluated each dialogue along four dimensions using a 4-point Likert scale (1=bad, 2=fair, 3=good, 4=excellent): \textit{correctness} of guidance, \textit{helpfulness} of assistance, \textit{alignment} with video content, and \textit{naturalness} of dialogue (detailed rubrics in Appendix). The evaluation achieved a weighted inter-rater agreement of 81\%, indicating strong consensus. As shown in Table \ref{table:human_eval_dialog} (top), the synthetic dialogues demonstrate consistently high quality, with average scores exceeding 3 across all dimensions. Notably, dialogue quality correlates with user interaction frequency, with more interactive dialogues scoring higher, particularly in naturalness. 

\noindent\textbf{Direct Comparison to Human Dialogues.} To contextualize these results, we additionally evaluated human dialogues from HoloAssist and WTaG on the same samples, enabling direct comparison between synthetic and human dialogues. Table \ref{table:human_eval_dialog} (middle and bottom) shows that \dataset's synthetic dialogues match or outperform their human-collected counterparts across multiple dimensions. This advantage is particularly notable in the HoloAssist subset, where our generated dialogues achieve significantly higher scores in helpfulness, correctness and naturalness. Qualitative analysis reveals that human-collected dialogues often contain artifacts from Wizard-of-Oz collection setups, where untrained individuals acting as assistants may not maintain consistent professional standards. In contrast, \dataset dialogues are designed to emulate standardized, professional assistant interactions, resulting in more consistent and helpful guidance. These results validate the effectiveness of our data curation pipeline in producing high-quality synthetic dialogues.



\begin{table}[t!]
\begin{minipage}{0.27\textwidth}
\centering
\resizebox{\columnwidth}{!}{
\addtolength{\tabcolsep}{-0.3em}
\begin{tabular}{lcc}
\toprule
\multicolumn{1}{c}{Metric} & P & S\\
\midrule
~~~~~~~~F1 vs Human     & 0.35\textsuperscript{**} & 0.32\textsuperscript{*}~ \\
Overall vs Human & 0.47\textsuperscript{**} & 0.44\textsuperscript{**} \\
Overall vs F1    & 0.67\textsuperscript{**} & 0.64\textsuperscript{**} \\
\bottomrule
\end{tabular}}
\vspace{-5pt}
\caption{Pearson and Spearman coefficient between our metrics and human judgment (*: $p < 0.05$, **: $p < 0.01$).\vspace{-12pt}}
\label{tab:human_corr}
\end{minipage}
~
\hspace{1pt}
~
\begin{minipage}{0.18\textwidth}
\centering
\resizebox{\columnwidth}{!}{
\addtolength{\tabcolsep}{-0.3em}
\begin{tabular}{ccc}
\toprule
\multicolumn{1}{c}{Metric} & A.N. &  D.G. \\
\midrule
F1     & 0.80 & 0.67 \\
Precision & 0.53  & 0.42 \\
Recall    & 0.47  & 0.63 \\
\bottomrule
\end{tabular}}
\vspace{-5pt}
\caption{Match rate between human and metric-based selection of the best $\theta$.\vspace{-12pt}}
\label{tab:human_pick}
\end{minipage}
\end{table}

    

\begin{table*}[t]
    \centering
    \resizebox{\textwidth}{!}{
    \addtolength{\tabcolsep}{-0em}
    \begin{tabular}{l|ccc|ccccccc}
    \toprule
    \multirow{2}{*}{Model}& \multicolumn{3}{c|}{Action Narration} & \multicolumn{7}{c}{Dialogue Generation} \\
      &    Precision &   Recall &   F1 &    Precision &   Recall &   F1  & Correctness & 
Promptness &      
Efficiency &      
Overall \\
    \midrule
     I=1        & 43.61 & 61.86 & 51.16 & 51.26 & 24.72 & 32.55 & 2.15 & 2.47 & 2.11 & 2.11 \\
     I=1 (w/ klg)  & - & - & - & 49.57 & 28.58 & 35.43 & 2.46 & 2.78 & 2.31 & 2.36 \\
     \hline
     I=5        & 62.81 & 61.12 & 61.96 & 44.41 & 26.36 & 32.62 & 2.13 & 2.46 & 2.09 & 2.10 \\
     I=5 (w/ klg)  & - & - & - & 44.24 & 31.52 & 36.25 & 2.50 & 2.78 & 2.34 & 2.41 \\
     \hline
     I=10      & 66.17 & 65.08 & 65.62 & 36.97 & 30.04 & 32.77 & 2.19 & 2.50 & 2.11 & 2.15 \\
     I=10 (w/ klg) & - & - & - & 37.54 & 34.93 & 36.07 & 2.53 & 2.83 & 2.31 & 2.42 \\
    \bottomrule
    \end{tabular}}
    \vspace{-10pt}
    \caption{Model evaluation results. Comparisons can be made across different tasks (Action Narration vs. Dialogue Generation), model variants (I=1, 5, or 10), and knowledge access setups (w/ or w/o knowledge). \vspace{-12pt}}
    \label{tab:main-result}
\end{table*}


\begin{table}[h]
    \centering
    \resizebox{\columnwidth}{!}{
    \begin{tabular}{l|ccc|ccc}
    \toprule
    \multirow{2}{*}{Model}& \multicolumn{3}{c|}{Action Narration}  & \multicolumn{3}{c}{Dialogue Generation} \\
      &   Precision &   Recall &   F1 &   Precision &   Recall &   F1 \\
    \midrule
     Baseline       &        20.5 &     56.6 & 30.1 &        49.6 &     25.9 & 32.9 \\
     $\rho$=0.2       &        56.4 &     40.6 & 47.2 &        48.5 &     26.7 & 33.5 \\
     $\rho$=0.1       &        51.3 &     68.6 & \textbf{58.7} &        48.0 &     27.5 & \textbf{34.4} \\
     $\rho$=0.01      &        58.5 &     52.9 & 55.5 &        35.9 &     33.0 & 34.2 \\
    \bottomrule
    \end{tabular}}
    \vspace{-5pt}
    \caption{Improvement from negative frame sub-sampling under different sub-sampling ratios.\vspace{-5pt}}
    \label{tab:w2t}
\end{table}
\begin{table}[t]
    \centering
    \begin{tabular}{l|ccc}
    \toprule
     Methods         &   Precision &   Recall &   F1 \\
    \midrule
    Drop-Middle     & 30.4 & 25.9 & 25.7  \\
    IPS (Ours)            & 49.6 & 25.9 & \textbf{32.9} \\
    \bottomrule
    \end{tabular}
    \vspace{-5pt}
    \caption{Comparison between inference-time context management method for long video processing.\vspace{-13pt}}
    \label{tab:summarization}
\end{table}

\begin{table}[t]
\centering
\resizebox{\columnwidth}{!}{
\addtolength{\tabcolsep}{-0.3em}
\begin{tabular}{l|cccc}
\toprule
 Subset&    Correctness & Promptness & Efficiency & Overall \\
\midrule
Ego4D        & 2.07 & 2.32 & 2.06 & 2.02 \\
HoloAssist    & 2.13 & 2.55 & 2.13 & 2.08 \\
EgoExoLearn   & 1.90 & 2.27 & 1.95 & 1.93 \\
Assembly101   & 1.94 & 2.24 & 1.98 & 1.93 \\
EpicKitchens & 2.07 & 2.26 & 2.04 & 2.02 \\
WTaG         & 2.79 & 3.16 & 2.51 & 2.67 \\

\bottomrule
\end{tabular}}
\vspace{-10pt}
\caption{Per-subset performance across domains.\vspace{-15pt}}
\label{tab:subset_result}
\end{table}


\vspace{-5pt}
\subsection{Validation of Proposed Metrics}
\label{sec:human_eval_metrics}

To measure whether our proposed metrics align with human judgment for model assessment, we conducted two human evaluations.

\noindent\textbf{Correlation with human preference in model ranking.} We selected 50 random tasks and collected predictions from three model variants, differing in tokens per frame and access to ground-truth recipes. Annotators ranked these predictions from best to worst (allowing ties), for comparison with rankings from our pairwise F1 score and LLM overall helpfulness score. Table \ref{tab:human_corr} shows that both metrics correlate positively with human judgment, with LLM scoring showing stronger alignment. We note that these correlation scores match those of previous automatic dialogue evaluation metrics \cite{yeh2021comprehensive,zhang2021dynaeval}, despite our additional challenge of measuring response timing. These results establish a baseline for developing metrics with better human correlation.


\noindent\textbf{Validation of speaking threshold selection.} The speaking threshold $\theta$ is a crucial hyperparameter that controls the model's balance between conservative and impulsive talking styles. To validate our use of validation F1 score for selecting $\theta$, we compared human preferences across different thresholds with our metric-based selections. The F1 score demonstrated the highest alignment with human preferences compared to precision and recall, achieving agreement rates of 0.8 for action narration (A.N.) and 0.67 for dialogue generation (D.G.), confirming its effectiveness as a selection criterion.


\vspace{-2pt}
\subsection{Result Analysis}
\vspace{-5pt}

We analyze our experimental findings to understand the challenges of proactive task guidance and evaluate the effectiveness of our proposed techniques.

\noindent\textbf{Limited gains from improved perception in dialogue generation.} Table \ref{tab:main-result} shows that while increasing tokens per frame ($I$) substantially improves action narration performance, it provides minimal benefits for dialogue generation. This indicates that effective task guidance requires more than just better visual perception. While the model becomes better at recognizing user actions, it still needs additional capabilities--such as long-horizon progress tracking, situational reasoning, and knowledge application--to provide meaningful guidance. These results highlight the fundamental challenges in our new problem formulation and emphasize the importance of higher-level reasoning capabilities.

\noindent\textbf{Benefits of task-specific knowledge.}
Table \ref{tab:main-result} also shows that providing the model with ground-truth knowledge (e.g., recipes) significantly improves guidance quality across all metrics. This improvement suggests that accessing recipes enables the model to align its guidance strategy with the specific plan shown in the video. This is critical for evaluation with pre-recorded demonstrations where multiple valid solutions exist but only one is shown. Without such knowledge, the model might be penalized for suggesting equally valid alternatives. We therefore recommend the knowledge-conditioned setup as the standard configuration for our evaluation framework. Becides, these findings also highlight the importance of retrieval-augmented generation (RAG) with task-relevant knowledge to improve real-world proactive assistant systems.

\noindent\textbf{Effectiveness of Negative Frame Sub-sampling (NFS).} We apply NFS with different sampling ratios $\rho$. As shown in Table \ref{tab:w2t}, training with NFS consistently improves the model's response timing decisions, with higher F1 scores across both tasks. The optimal performance is achieved at $\rho=0.1$, which we adopt for all subsequent experiments.

\noindent\textbf{Effectiveness of Iterative Progress Summarization (IPS).}
Direct ablation of IPS is infeasible, as the evaluation cannot complete for videos exceeding the model's training context length. We instead compare against a modified version of StreamingLLM \cite{xiaoefficient}--a context management approach that handles memory constraints by dropping middle tokens while preserving initial task goals. Table \ref{tab:summarization} shows that IPS significantly outperforms this baseline, demonstrating its effectiveness in long-term task progress tracking.  

\noindent\textbf{Task familiarity impacts performance.}
Table \ref{tab:subset_result} reveals significant performance variation across \dataset subsets. The model performs notably better on WTaG tasks, which contain only three unique tasks that appear in training (albeit in different environments during evaluation). In contrast, performance drops substantially for EgoExoLearn and Assembly101 tasks, due to relatively less training samples available for laboratory and assembly domains. These results highlight the need to improve generalization to new tasks and domains. 

\vspace{-5pt}
\section{Conclusion}
\vspace{-5pt}

We introduce a novel framework for perceptual task guidance through streaming video dialogue generation, supported by \dataset--a large-scale synthetic dataset, validated evaluation metrics, and an enhanced end-to-end model. Our experiments reveal that while visual perception alone has limited impact, task knowledge and effective memory mechanisms significantly improve performance. We hope the curated data, new evaluation metrics, and our baseline models will provide much needed  resources and insights, establishing a foundation for advancing real-time AI assistance.

\section*{Limitations}

Our dialogue synthesis pipeline, while carefully designed, has room for improvement in quality control. As shown in Table \ref{table:human_eval_dialog}, the alignment between dialogues and video content requires enhancement. Future work could leverage more advanced LLMs, refined prompt engineering, or incorporate multimodal models to increase synthesis quality.

The dataset's reliance on pre-existing video annotations limits its scalability, as such annotations are expensive and time-consuming to obtain. Recent advances in multimodal LLMs \cite{achiam2023gpt,TheC3,team2024gemini} open the possibility of generating dialogues directly from raw videos, which could make data synthesis more efficient and scalable.

While our automatic evaluation metrics show promise, their validation is limited to our current experimental setup. These metrics need broader testing across diverse models, performance levels, and related tasks. Additionally, our text-only evaluation approach could be enhanced by incorporating multimodal metrics that consider video content, to establish more robust benchmarks for interactive assistant systems.

Another limitation is that we do not explicitly model utterance duration that regards precise user-assistant turn-taking simulation. However, for the core task defined in our work, determining when and how to provide proactive guidance based on streaming video context, evaluating the timing of interventions is a starting point towards more elaborate timing/duration management in the future.

Finally, while our proposed proactive assistant model is the first to tackle this challenge, its performance remains suboptimal. As shown in Table \ref{tab:subset_result}, even on the best-performing domain, the model falls below acceptable thresholds in LLM evaluation (overall scores below 3 out of 5). Notably, it struggles with response timing, dialogue consistency, and delivering detailed guidance that demands fine-grained perception. These limitations underscore the need for better modeling of speaking time, stronger visual-language alignment, and enhanced visual understanding in streaming dialogue generation.

\section*{Ethics Statement}
All datasets used in this work are publicly available and do not contain sensitive or private information. The models used in our data synthesis pipeline and experiments are based on open-source frameworks and will be released upon publication. We acknowledge that LLM-generated utterances may exhibit hallucinations or biases, and we conduct human evaluations to understand the quality of our synthetic dialogue data.

\section*{Acknowledgments}
This work was supported by Meta and the DARPA PTG program HR00112220003.
We would like to thank the anonymous reviewers for their valuable comments and suggestions.

\bibliography{custom}

\appendix
\clearpage
\setcounter{page}{1}
\setcounter{section}{0}
\renewcommand*{\thesection}{\Alph{section}}

\lstset{
  basicstyle=\ttfamily\footnotesize,
  columns=flexible,
  frame=single,
  captionpos=b,
  breaklines=true,
  breakindent=0pt,
  breakatwhitespace=true,
  moredelim=[s][\bfseries\color{blue}]{[}{]}
}


\section{Dataset}


\subsection{Synthetic Dialogue Data Generation Details}

\begin{table*}[ht]
\centering
\resizebox{\textwidth}{!}{
\addtolength{\tabcolsep}{-0.1em}
\begin{tabular}{@{}lccccccl@{}}
\toprule
Dataset & \textbf{Domain} & \textbf{\#Tasks} & \textbf{\#Videos} & \textbf{Total Duration} & \textbf{Avg Duration} & \textbf{Labels} \\ \midrule
Ego4D-Goalstep \cite{grauman2022ego4d,song2024ego4d}) & Cooking & 86 & 851 & 368h & 26m & C + F \\
EpicKitchen \cite{damen2020epic,Damen2022RESCALING} & Cooking & - & 700 & 100h & 9m & F \\
HoloAssist \cite{wang2023holoassist} & Object Manipulation; Assembly & 20 & 2221 & 166h & 5m & C + F + M + D \\
EgoExoLearn \cite{huang2024egoexolearn} & Cooking; Laboratory Tasks & 8 & 432 & 96h & 13m & C + F + R \\
Assembly101 \cite{sener2022assembly101} & Assembly & 101* & 356 & 42h & 7m & C + F + M \\
WTaG \cite{bao2023can} & Cooking & 3 & 56 & 10h & 11m & C + M + D + R \\ \bottomrule
\end{tabular}}
\vspace{-5pt}
\caption{Summary of egocentric video datasets used in \dataset. The statistics presented are as originally reported in the corresponding papers before filtering. The number of tasks indicates the types of tasks in each dataset, except for Assembly101, where it represents the unique number of toys. Label abbreviations are as follows: C (coarse action labels), F (fine-grained action labels), M (mistake and correction labels), D (human-collected assistant-user dialogues), and R (ground-truth recipes).\vspace{-10pt}}
\label{tab:dataset_summary}
\end{table*}
\label{supp:data-syn}
Videos in \dataset are sourced from six extensively labeled datasets: Ego4D \cite{grauman2022ego4d} with GoalStep annotations \cite{song2024ego4d}, EpicKitchen \cite{damen2020epic,Damen2022RESCALING}, HoloAssist \cite{wang2023holoassist}, Assembly101 \cite{sener2022assembly101}, EgoExoLearn \cite{huang2024egoexolearn}, and WTaG \cite{bao2023can}. Detailed statistics and label types are summarized in Table \ref{tab:dataset_summary}. When generating timestamped video descriptions for these videos, we leverage all available labels to provide a comprehensive understanding of the video content. In cases where both coarse- and fine-grained action labels are available, they are organized into hierarchical formats for clarity. An example of this unified timestamped video description, incorporating coarse and fine-grained actions, mistake corrections, and assistant-user dialogues, is shown in List \ref{example_video_description}.

\begin{lstlisting}[basicstyle=\ttfamily\footnotesize,caption={An example video description from HoloAssist.},label={example_video_description},language={}]
[11.0s-28.2s] The user grabs the GoPro.
 - [11.0s-12.2s] approach gopro
 - [13.7s] assistant: "Okay."
 - [15.1s] assistant: "You can pull the GoPro."
 - [16.9s] user: "GoPro."
 - [17.0s-22.2s] grab gopro
 - [22.2s-23.5s] flip bag
[29.9s-66.2s] The user changes the battery for the GoPro.
 - [30.2s] assistant: "Change the Battery."
 - [34.2s-41.0s] pull battery door
 - [41.0s-42.3s] open battery door
 - [43.5s-45.0s] grab battery
 - [45.1s-46.1s] withdraw battery
 - [45.6s] assistant: "Take out the battery."
 - [47.5s-49.6s] place battery
 - [48.8s] assistant: "Now, put it down"
 - [49.7s-50.9s] lift battery
 - [50.9s-51.7s] insert battery
 - [58.8s] assistant: "Close it."
 - [59.8s-61.0s] close battery door
 - [62.2s-63.5s] push battery door
 - [67.0s] assistant: "Now change the micro SD."
[68.4s-304.3s] The user opens the GoPro.
 - [68.6s-70.8s] grab battery door
 - [70.8s-72.7s] open battery door
 - [72.7s-73.9s] press battery (ERROR: The user presses the wrong place.)
 - [73.9s] assistant: "No that one."
 ...
\end{lstlisting}

Next we provide additional details for each step in our data curation pipeline. 

\paragraph{Task Goal and Recipe Generation} The objective is to generate a high-level task goal and recipe-style instructions that outline the key steps required to complete the task, derived from the video descriptions. For datasets where these elements are already provided in their labels (EgoExoLearn, WTaG), this step is skipped. During generation, we observe that the LLM can sometimes be distracted by irrelevant actions in the video descriptions, resulting in less accurate or unstable outputs across different sampling trials. To address this issue, we employ a two-step process. First, we generate 10 candidate recipes\footnote{We use vLLM\cite{kwon2023efficient} for efficient parallel sampling.} using the prompt shown in List \ref{prompt:task_infer}. Next, we refine these recipes into a single cohesive and integrated version by calling the LLM one more time with the prompt in List \ref{prompt:task_refine}. 

\begin{lstlisting}[caption={Prompt for inferring task goal and recipe from video descriptions.},label={prompt:task_infer},language={},moredelim={[s][\bfseries\color{purple}]{\{}{\}}}]
Here is a video description of an experienced user working on the task - {goal_description}:
{video_descriptions}

Try to infer the **high-level** recipe from the descriptions. Note that the steps may not belong to the same trial, so you have to infer the correct order of the steps based on common sense, and re-order the steps if necessary. Do not hallucinate details that are not mentioned in the descriptions. Also generate a more **informative** and **descriptive** name for the task based on provided descriptions. The name should be a description of the task, instead of the name of the recipe. 

Give plain and concise text with numbered key steps in the following format:
[task name]  1. ...   2. ...
\end{lstlisting}
\begin{lstlisting}[caption={Prompt for task goal and recipe refinement.},label={prompt:task_refine},language={},moredelim={[s][\bfseries\color{purple}]{\{}{\}}}]
Here are {num_repeats} {knowledge_type}s:
{recipes}

Some may be incorrect or incomplete. Please give a single correct and complete {knowledge_type} for the task, with numbered key steps. Pick the title that is descriptive for the task, instead of a {knowledge_type} name.

Give plain, unformatted and concise text with numbered key steps in the following format:

[task name]  1. ...  2. ...

Do not include any other information or note.
\end{lstlisting}

\paragraph{Video Pre-Filtering} 
The next step is to filter out videos that are unsuitable for proactive assistant-user dialogue modeling. First, we exclude videos with low label coverage, as their descriptions may lack sufficient detail to provide a clear understanding of the content. Then, using the video descriptions, task goals, and recipes generated in the previous step, along with the domain and recipe type derived from the dataset metadata, we prompt the LLM to classify each video into one of three categories:
\begin{itemize}
    \item 0: The task does not belong to the target domain.
    \item 1: The camera wearer performs the target task following the recipe.
    \item 2: The camera wearer performs other tasks simultaneously while working on the target task.
\end{itemize}
We keep only the videos classified as category 1 for subsequent steps. To minimize noise, this classification process is repeated 10 times for each task, and the majority label is used as the final classification.

\begin{lstlisting}[caption={Prompt for video pre-filtering.},label={prompt:video_label},language={},moredelim={[s][\bfseries\color{purple}]{\{}{\}}}]
Here is a video description of an user working on the task - {goal_description}:
{video_descriptions}

Reference {knowledge_type}:
{knowledge}

Is this a {knowledge_type}? If so, was the user likely to:
1. perform the task roughly following the {knowledge_type} (**no** need to be strict), OR
2. perform other tasks (or another trial of the same task) simultaneously in a multi-tasking manner?

Answer with your analysis, and end your response with "Final answer: 1, 2 or 0" (0 denotes that the activity is not related to {domain}).
\end{lstlisting}

\paragraph{Multi-Round Dialog Generation} To simulate realistic dialogues aligned with video content, we design a detailed instruction prompt, as shown in List \ref{prompt:dialog_gen}. We incorporate dataset-specific instructions to account for variations across data sources to improve generation quality (List \ref{prompt:dataset_specific}). To simulate diverse user behaviors, we define three type of user profiles that provide high-level guidelines for user interaction:
\begin{itemize}
    \item \texttt{no\_talk}: The user follows the assistant's instructions without speaking.
    \item \texttt{talk\_some}: The user occasionally asks questions or seeks confirmation about instructions, accounting for approximately 20\% of the steps.
    \item \texttt{talk\_more}: The user is talkative, asking both task-related and unrelated questions, accounting for approximately 40\% of the steps.
\end{itemize}
Since we can sample different user behaviors for the same video, our synthetic dataset can be easily expanded by generating multiple variations. Specifically, we create 10 dialogues per video, distributed across user types in a 2:4:4 ratio. In practice, we observe that very long videos can result in excessively lengthy prompts, which may lead to poor alignment between the video description and the generated dialogue due to performance degradation of LLMs when processing long contexts \cite{liu2024lost}. To address this, we propose an iterative approach to generate dialogues within a limited time window chunk by chunk. For each chunk, we provide only the video description corresponding to that time window and up to the 10 most recent dialogue turns to ensure contextual consistency. This modification significantly improves alignment between the video description and the generated dialogue while stabilizing memory consumption due to the reduced prompt length. After generation, we conduct an additional refinement step to enhance dialogue quality. Specifically, we prompt the LLM to merge dialogue turns that occur close in time, improve naturalness and fluency by incorporating more coreference and pronouns, make assistant responses more concise, and avoid unfriendly behaviors. Refer to List \ref{prompt:dialog_refinement} for details.

\begin{lstlisting}[caption={Prompt for dialogue simulation.},label={prompt:dialog_gen},language={},moredelim={[s][\bfseries\color{purple}]{\{}{\}}}]
Here is a video description of an user working on the task - {goal_description}:
{video_descriptions}

Your goal is to simulate a conversation between the user and an assistant, where the user's actions are performed following the assistant's instructions. The user will first mention the overall goal of the task. The assistant informs the user about the next step at proper time. Importantly, the assistant is proactive and always provides the next step even before the user asks for it. Before the task starts, the assistant may also give a brief introduction about the task. {additional_requirement}

Requirements for the assistant:
- Time is crucial! Try to generate the dialog that strictly aligns with the video timeline.
- Try to cover all the essential steps in the task. If the user asks a question at the time the assistant should give the next step, the assistant turn should include both the response to the question and instruction about the next step.
- Be helpful and friendly. If the user asks something that has been explained before, the assistant should still provide the information with patience.
- Try to be encouraging when the user makes progress, but do not overdo it.
- Be concise! The dialog is verbal, so avoid long sentences.
- Do not say "can you do it for me" to the user.


Requirements for the user:
{user_requirement}


Generation format:
[time] User: ...
[time] Assistant: ...
[time] Assistant: ...
[time] User: ...
[time] Assistant: ...

Note that the minimal interval between each turn is 1 second, which means the user will wait for at least 1 second after an assistant's turn, and two consecutive assistant's turns should have at least 1 second interval. Combine close turns into a single turn if necessary. One exception is that the assistant must respond **immediately** when the user says something (i.e. give a response right after an user's turn at the same time).

{dialog_history}

In this round, please **only** generate the dialog for the video from time [{start_time:.1f}s] to [{end_time:.1f}s]!
\end{lstlisting}

\begin{lstlisting}[caption={Prompt specific for each dataset as additional requirements.},label={prompt:dataset_specific},language={},moredelim={[s][\bfseries\color{purple}]{\{}{\}}},morekeywords={HoloAssist,EgoExoLearn,Epickitchens,WTaG,Assembly101},keywordstyle={\textbf}]
HoloAssist: Note that the video description contains both the user's actions and the user-assistant dialog. Anchor the simulated dialog to the existing dialog, and try to rephrase the utterances to make them more coherent and human-like. You may add a few more turns around the **essential steps** of the task, which are the underlying intentions of the action instead of the actions themselves. Add a few turns to make the dialog more fluent and helpful, but avoid being overwhelming.

EgoExoLearn: The simulated dialog should be centered around the **key steps** of the task, not every single action of the user. Try to make the dialog more coherent and helpful as what a human assistant will say.

Epickitchens: The simulated dialog should be centered around the **key steps** of the task, not every single action of the user. Note that the user may make mistake or perform suboptimal actions, the assistant should not give instructions on those actions, but smartly select right time to give guidance. Try to make the dialog more coherent and helpful as what a human assistant will say.

WTaG: Note that the video description contains both the step description and the user-assistant dialog. Anchor the simulated dialog to the existing dialog, and try to rephrase the utterances to make them more coherent and human-like. Add more details such as assistant feedback or user question during long steps if necessary. Remember to generate the response to user's question even if there isn't one in the original dialog from the video description.

Assembly101: The mistakes made by the user are marked by (mistake: <mistake type>). If a mistake happens, we want to simulate the dialog in the way that the assistant helps the user correct the mistake. To be more specific, the assistant SHOULD NOT give instructions if an action is 'wrong order', 'previous one is mistake' or 'shouldn't have happened'. Instead, the assistant should give instruction of the CORRECT next step (i.e. scan the future actions and select the nearest correct action). Afterwards, at the start of actions marked as 'correction', the assistant should mention the previous mistake and give insruction on how to correct it based on the corrective action. For 'wrong position' mistakes, the assistant can give the instruction of that action, but need to point out the mistake at the start time of corrective action for that mistake.
\end{lstlisting}

\begin{lstlisting}[caption={Prompt for dialogue refinement and intent labeling for assistant turns.},label={prompt:dialog_refinement},language={},moredelim={[s][\bfseries\color{purple}]{\{}{\}}}]
Here is a conversation between a user and an assistant:
{dialog_history}

For each assistant message, add labels regarding the assistant's initiativity and intention:

Initiativity:
- initiative: The assistant says something proactively without the user asking for it.
- responsive: The assistant responds to the user's question or comment.

Intention:
- instruction: The assistant gives an instruction to the user.
- correction: The assistant corrects a mistake made by the user, either proactively or responsively. Suggestions for alternative actions can also be included.
- info_sharing: The assistant shares some information with the user, such as explaining something or giving a tip.
- feedback: The assistant gives feedback to the user, such as "good job" or "tips for improvement".
- other: Other intentions that do not fall into the above categories.

Intention can be multiple, e.g., "instruction, info_sharing".

Generation format:
[time] User: ...
[time] Assistant: ... [initiativity|intentions]
[time] Assistant: ... [initiativity|intentions]
[time] User: ...
[time] Assistant: ... [initiativity|intentions]

When generating the dialog, you should also refine the dialogue following these guidelines:
1. Merge turns that are close in time (less than 1 second apart) into a single turn, when the content is similar or related.
2. Use more coreference and pronouns to make the dialog more coherent and human-like.
3. Decide the length of assistant messages smartly. Make them more clear and helpful when necessary, but keep them concise and to the point in general.
4. Avoid repeating the same talking patterns or phrases. For example, do not say "make sure ..." for every instruction.
5. Rephrase impolite or inappropriate language, such as "as I have mentioned this earlier ...", to be more friendly and helpful. But keep concise and to the point.
6. Remove anything other than the dialog itself, such as the user's actions or explanations of how the dialog is generated.
Do not just copy paste the original dialog!
\end{lstlisting}

\paragraph{Dialogue Annotation}
To facilitate the analysis of our generated dialogues, we use LLM to annotate the initiativity (responsive or initiative) and intention type (instruction, correction, information sharing, feedback, and other) of each assistant turn. We find such annotation can be effectively generated within the dialogue refinement step, using a single LLM call with the prompt in List \ref{prompt:dialog_refinement}. Additionally, we generate progress summaries at each assistant turn to support the iterative progress summarization approach (\S\ref{sec:summarization}. These summaries include details such as the elapsed time, the task goal mentioned by the user, completed steps as progress, topics discussed by the user, and the current state or step of the task (List \ref{prompt:summarize}). 

\begin{lstlisting}[caption={Prompt for progress summary generation},label={prompt:summarize},language={},moredelim={[s][\bfseries\color{purple}]{\{}{\}}}]
Here is a conversation between a user and an assistant:
{dialog_history}

Summarize the task goal and progress so far, including:
1. The task goal mentioned by the user.
2. What has been done.
3. Other topics mentioned by the user in the conversation, if any.
4. The current state/step of the task.
Be faithful and try to include all the relevant information.

Give your response in plain text of a single line in the following format:
SUMMARY: <progress summary>
\end{lstlisting}

\paragraph{Automatic Quality Evaluation}
To evaluate the quality of the generated dialogues, we assess their alignment and step coverage with the corresponding video descriptions. We first extract all time steps from the video descriptions, denoted as $T_v$, and all time steps from the generated dialogues, denoted as $T_d$. For each time step in $T_d$, we identify its closest time step in $T_v$ and compute the average time difference across all pairs, normalized by the number of dialogue turns. Similarly, for each time step in $T_v$, we find its closest match in $T_d$ and calculate the average time difference, normalized by the number of video description steps. These values approximate the precision of dialogue turns relative to the video (p) and the recall of task steps in the video descriptions (r). Additionally, to ensure the assistant remains responsive, we count the number of user turns without an immediate assistant response as a penalty term $nr$. The final quality score is computed as $score = 10 - p - r - nr$, the higher the better.

\paragraph{Post-Filtering and Data Splitting}
We derive our training set from the training splits provided by each original dataset, while our validation and testing sets are based on the respective validation splits. For the training set, we filter out dialogues with a score below 3. For the validation sets, we retain only the highest-scoring dialogue for each user type. If any dialogue for a video scores below 5, the video is removed. From the remaining videos, where each has three dialogues, we evenly split them into validation and test sets. This process removes approximately 25\% of the dialogues and 41 hours of video.

\begin{figure*}[t]
\centering
\begin{subfigure}[b]{0.30\textwidth}
    \centering
    \includegraphics[width=\textwidth]{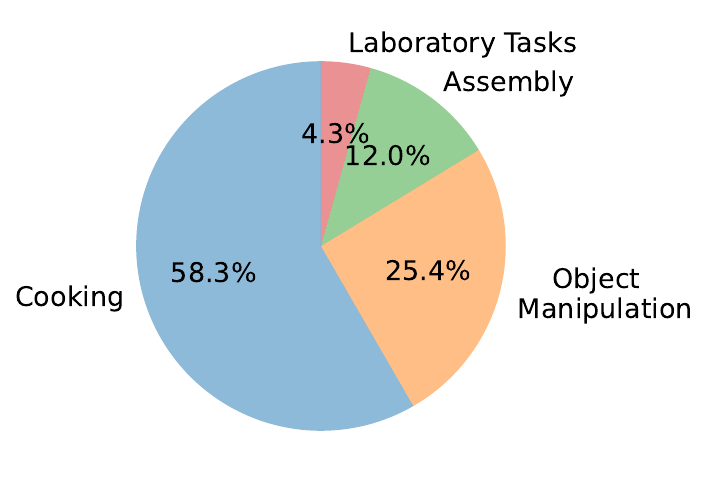}
    \caption{Task Domain Distribution.}
    \label{fig:domain_dist}
\end{subfigure}
~
\begin{subfigure}[b]{0.32\textwidth}
    \centering
    \includegraphics[width=\textwidth]{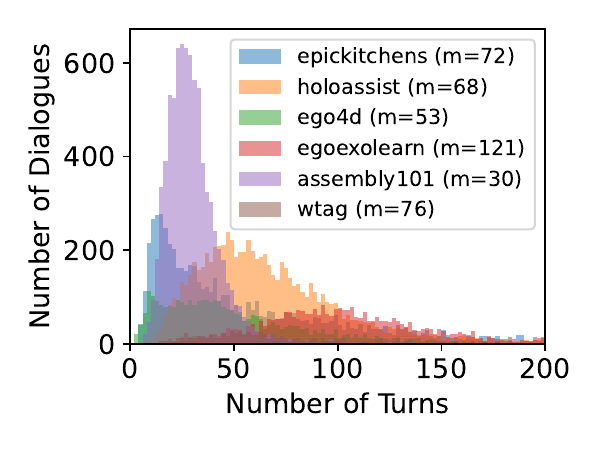}
    \caption{Dialogue length distribution.}
    \label{fig:turn_num_dist}
\end{subfigure}
~
\begin{subfigure}[b]{0.32\textwidth}
    \centering
    \includegraphics[width=\textwidth]{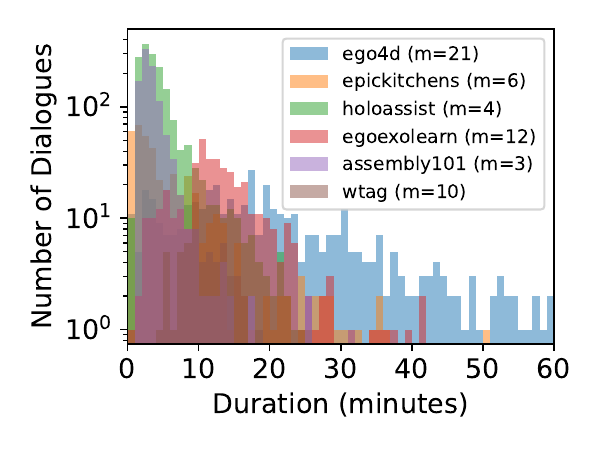}
    \caption{Task duration distribution.}
    \label{fig:duration_dist}
\end{subfigure}
~
\begin{subfigure}[b]{0.32\textwidth}
    \centering
    \includegraphics[width=\textwidth]{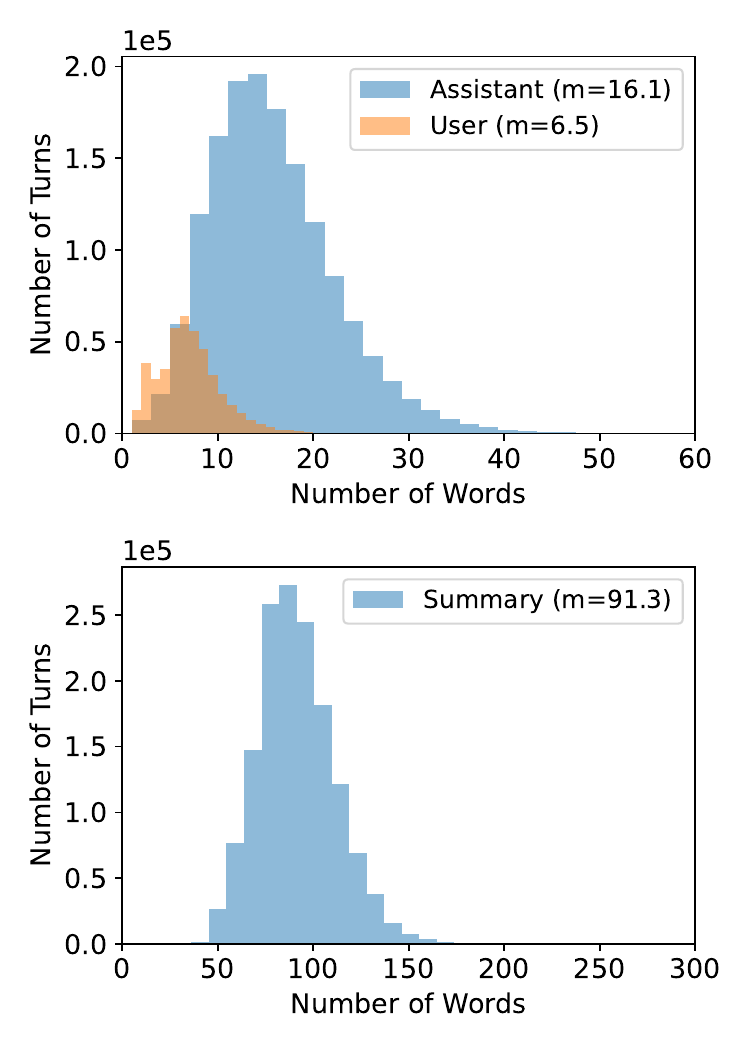}
    \caption{Length distribution for assistant, user utterances and our generated progress summary.}
    \label{fig:length_dist}
\end{subfigure}
~
\begin{subfigure}[b]{0.65\textwidth}
    \centering
    \includegraphics[width=\textwidth]{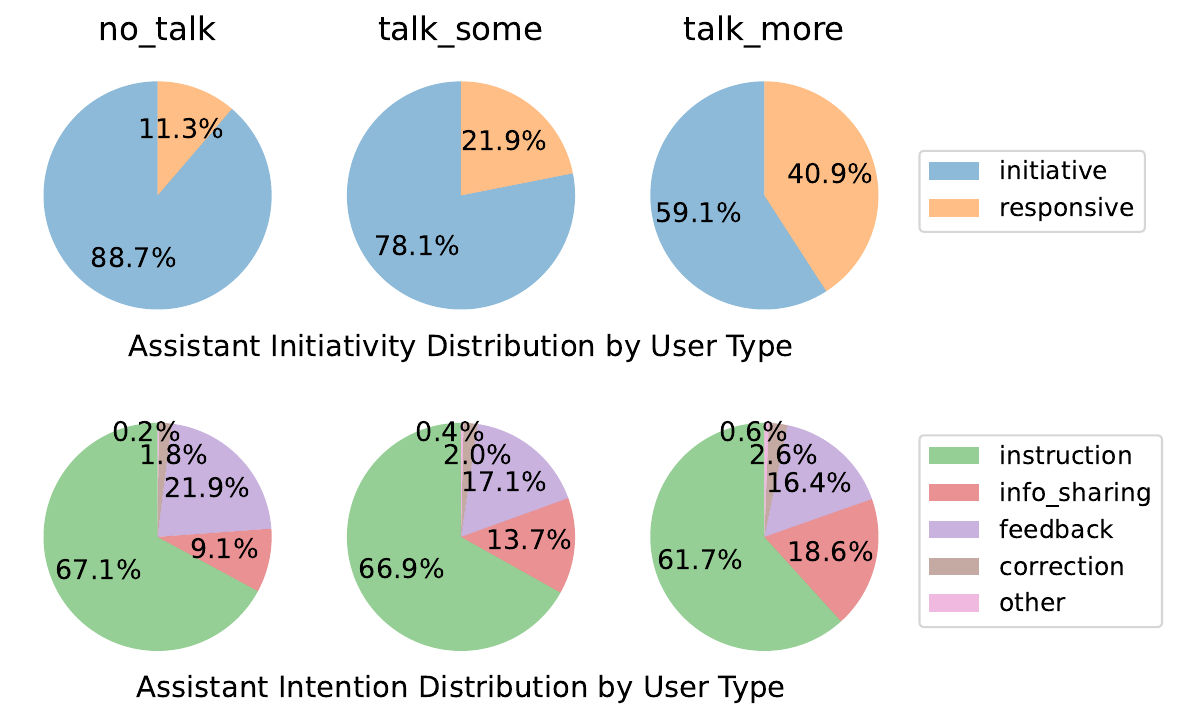}
    \vspace{3pt}
    \caption{Assistant initiativity and intention distribution by user type.}
    \label{fig:assistant_turn_dist}
\end{subfigure}
\caption{\dataset dataset statistics overview including task domain distribution, dialogue structure, duration variability, and assistant interaction patterns. Mean values for each distribution are provided in the legend.\vspace{-7pt}}
\label{fig:statistics}
\end{figure*}


\subsection{Implementation Details}
We utilize LLaMA-3.1-70B-Instruct \cite{dubey2024llama} as the LLM for all the steps described above. The model is hosted locally using vLLM \cite{kwon2023efficient}, running a FP8-quantized version\footnote{\url{https://huggingface.co/neuralmagic/Meta-Llama-3.1-70B-Instruct-FP8-dynamic}} on four H100 GPUs. Although we use a specific LLM for data generation, our pipeline is model-agnostic and can be readily adapted to more advanced models with minimal prompt modifications. We hope our open-sourced prompt designs will support future efforts in curating higher-quality datasets with more capable LLMs. 

\subsection{Data Statistics and Distributions}
\label{supp:statistics}
Figure \ref{fig:statistics} provides a comprehensive overview of the \dataset data statistics. Specifically, Figure \ref{fig:domain_dist} presents the distribution of task domains by video durations, showing that the dataset predominantly contains cooking tasks (58.3\%), followed by object manipulation (25.4\%), assembly (12.0\%), and laboratory tasks (4.3\%). Figure \ref{fig:turn_num_dist} illustrates the dialogue length distribution, measured by the number of turns per dialogue, highlighting a significant variability with some dialogues exceeding 200 turns. Figure \ref{fig:duration_dist} shows the task duration distribution where the majority of tasks last under 20 minutes, while some extend up to an hour. Figure \ref{fig:length_dist} visualizes the length distribution of user and assistant utterances, as well as the generated progress summaries. Assistant utterances are generally longer (mean = 16.1 words) compared to user utterances (mean = 6.5 words). For the generated summaries, the average length is 91.3 words, with the largest summary less than 200 words, showing that the information can be successfully compressed into such summarizes without growing linearly with the dialogue. Figure \ref{fig:assistant_turn_dist} highlights assistant initiativity and intention distributions across three user types. As users speak more frequently, the assistant responds with more reactive utterances as expected. Regarding assistant intentions, the majority of utterances provide instructions (over 60\%), with information sharing and feedback also being common. Mistake corrections occur less frequently, primarily because most of the videos used are error-free, underscoring the need to collect more task execution videos that include mistakes.

\subsection{Examples}
\begin{figure*}[t]
    \centering
    \includegraphics[width=0.84\linewidth]{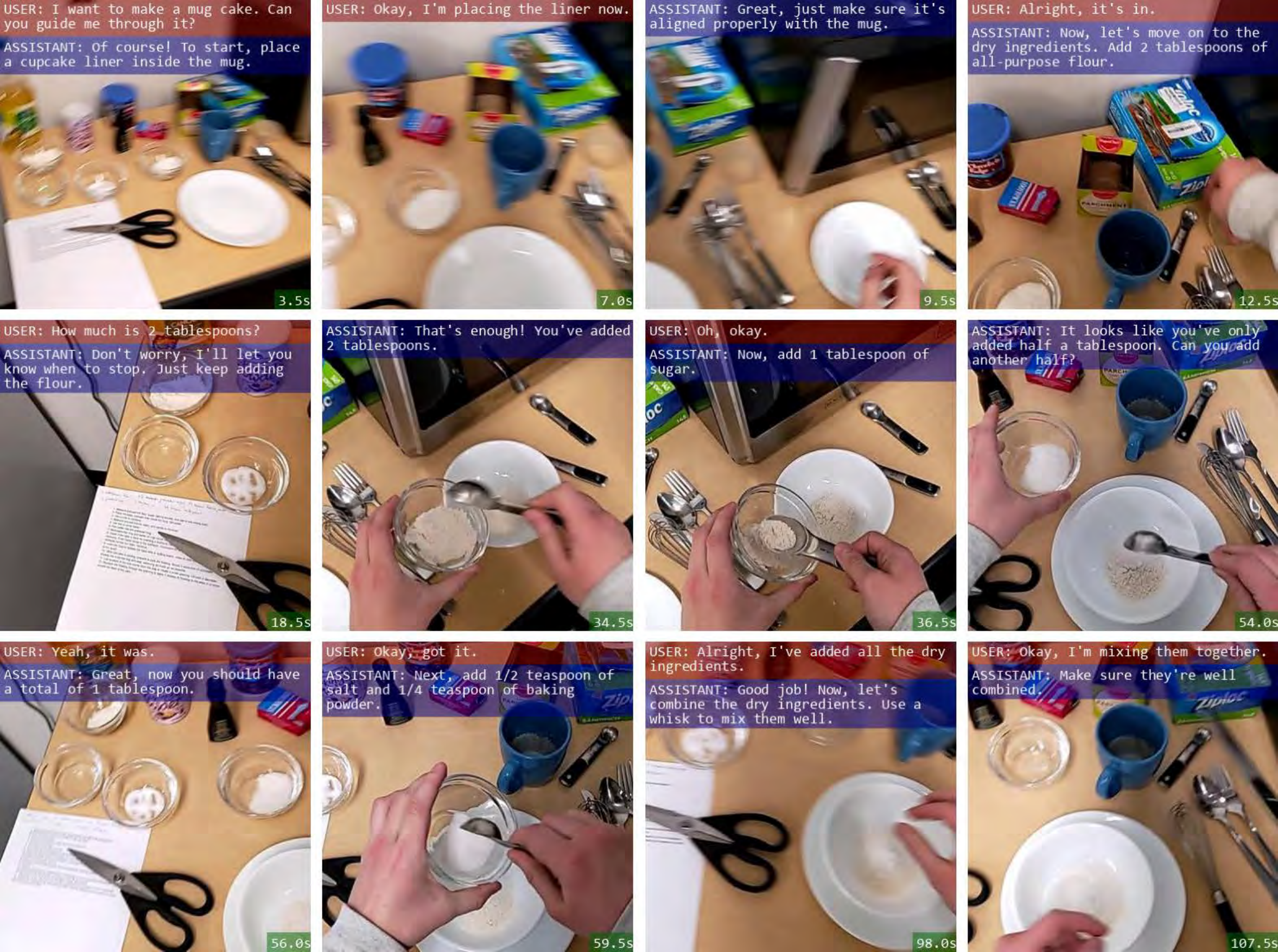}
    \vspace{-8pt}
    \caption{Example of a cooking task in \dataset. \vspace{-8pt}}
    \label{fig:data-sample1}
\end{figure*}

\begin{figure*}[t]
    \centering
    \includegraphics[width=0.84\linewidth]{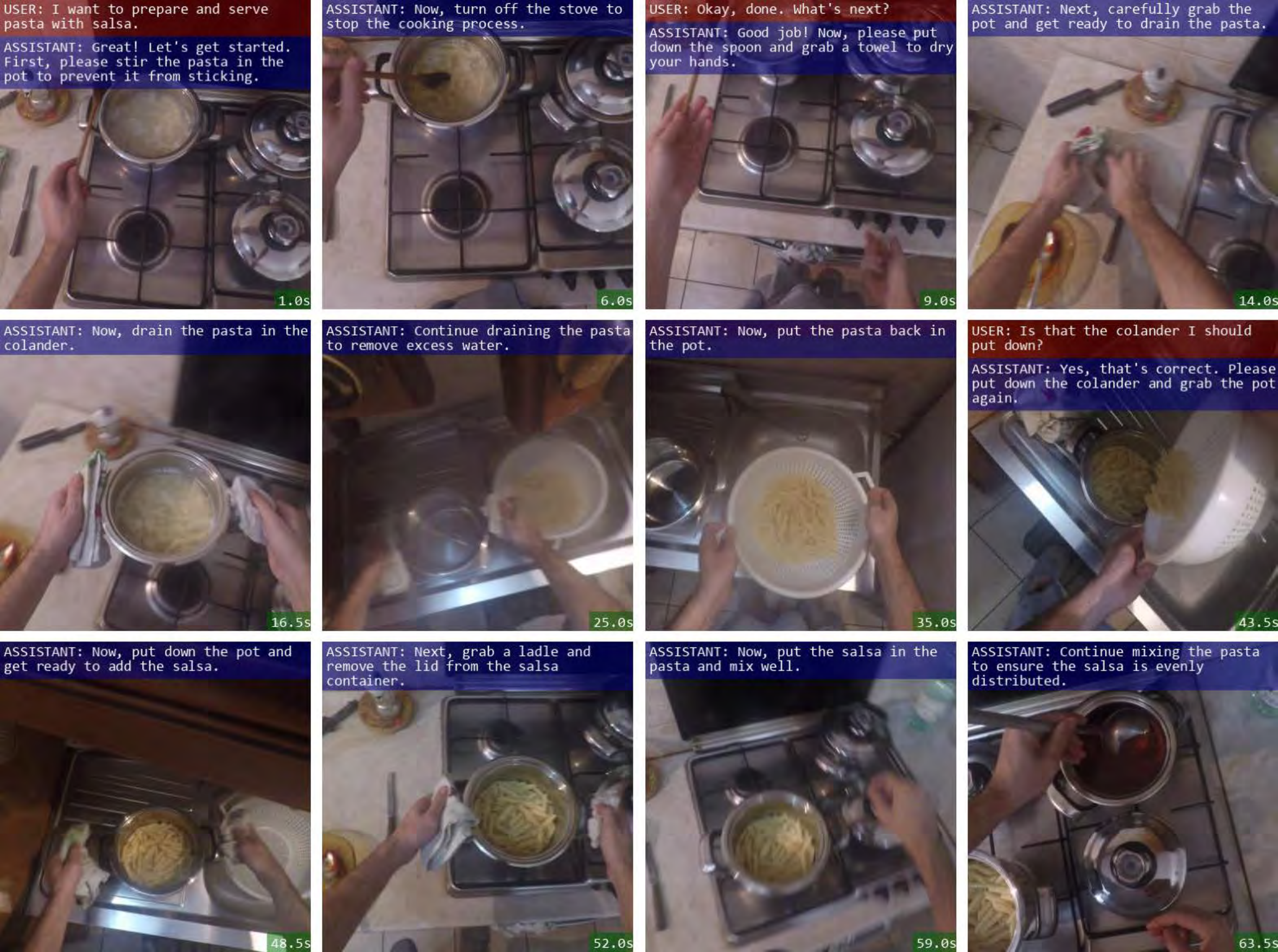}
    \vspace{-8pt}
    \caption{Example of a cooking task in \dataset.\vspace{-8pt}}
    \label{fig:data-sample2}
\end{figure*}

\begin{figure*}[t]
    \centering
    \includegraphics[width=0.84\linewidth]{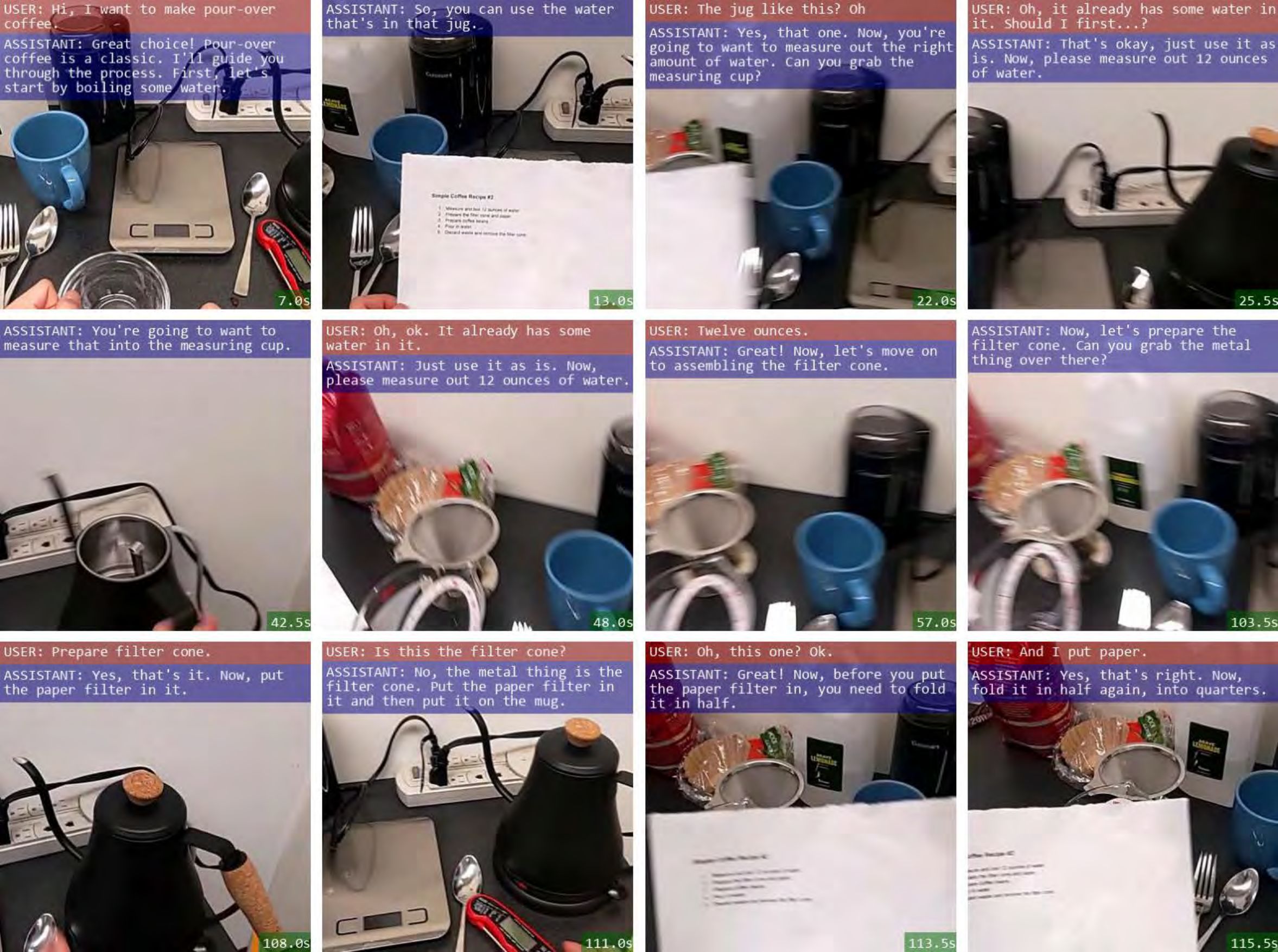}
    \vspace{-8pt}
    \caption{Example of a cooking task in \dataset.\vspace{-8pt}}
    \label{fig:data-sample3}
\end{figure*}

\begin{figure*}[t]
    \centering
    \includegraphics[width=0.84\linewidth]{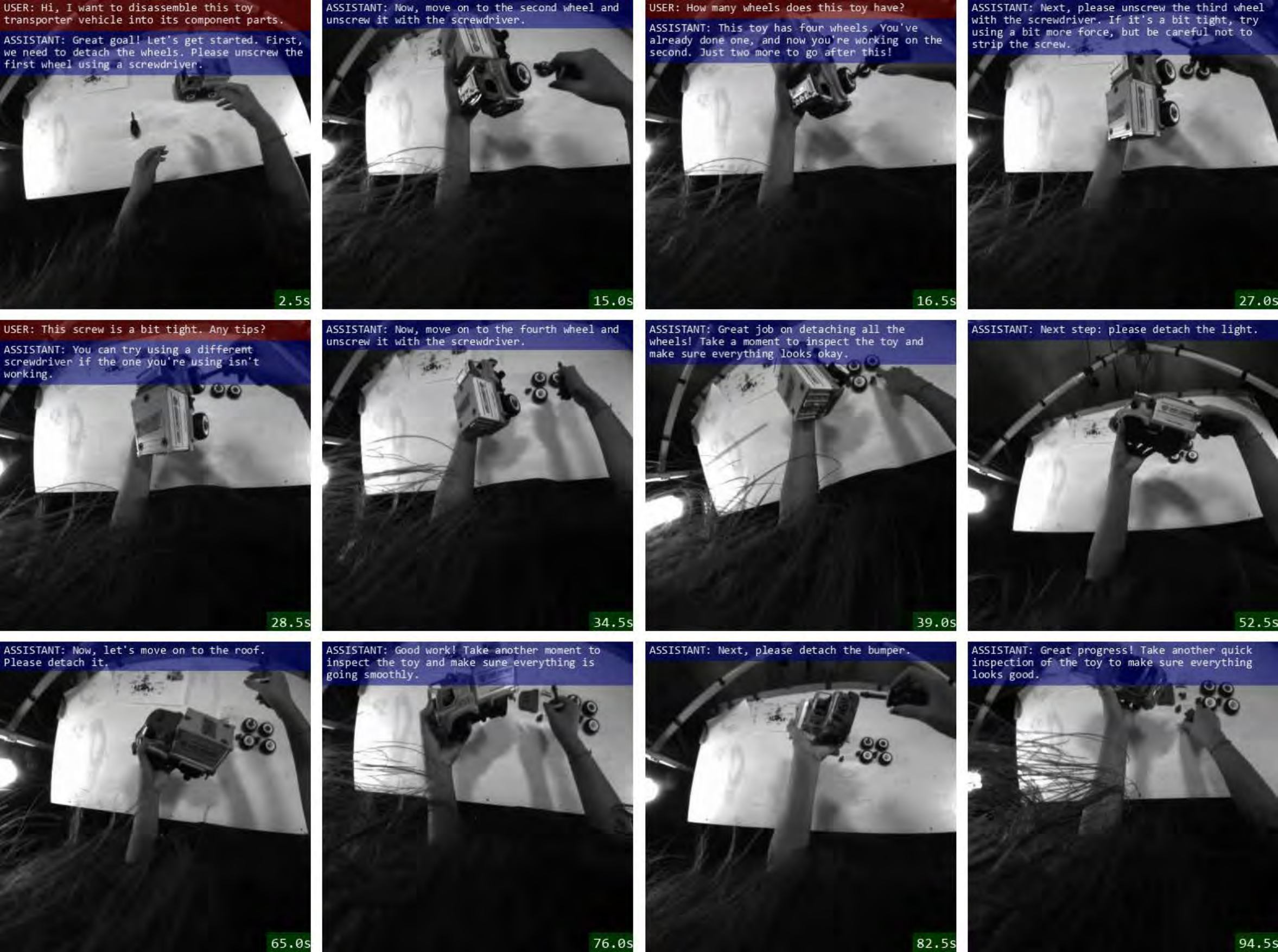}
    \vspace{-8pt}
    \caption{Example of a assembly task in \dataset.\vspace{-8pt}}
    \label{fig:data-sample4}
\end{figure*}

\begin{figure*}[t]
    \centering
    \includegraphics[width=0.84\linewidth]{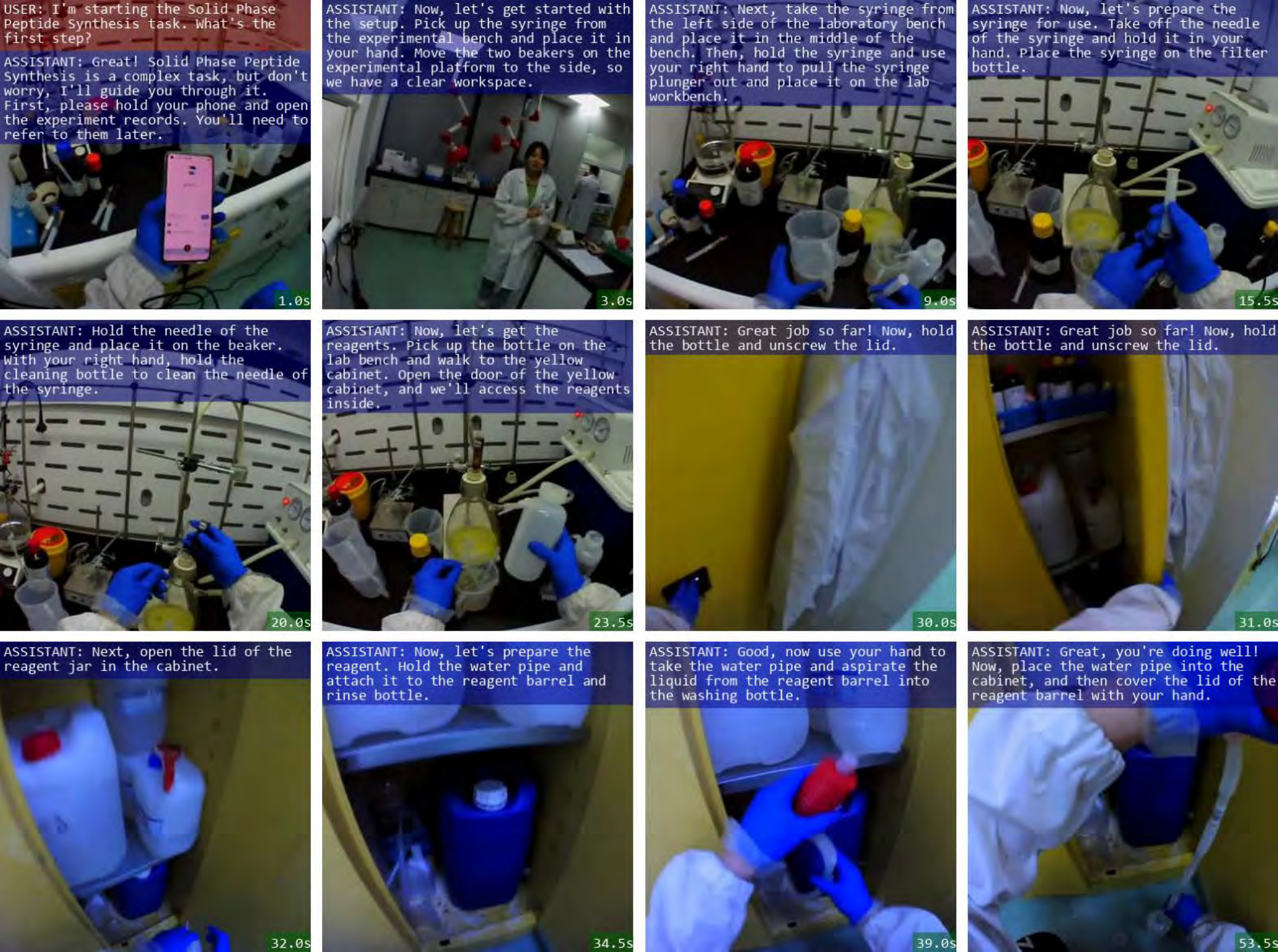}
    \vspace{-8pt}
    \caption{Example of a laboratory task in \dataset.\vspace{-8pt}}
    \label{fig:data-sample5}
\end{figure*}

\begin{figure*}[t]
    \centering
    \includegraphics[width=0.84\linewidth]{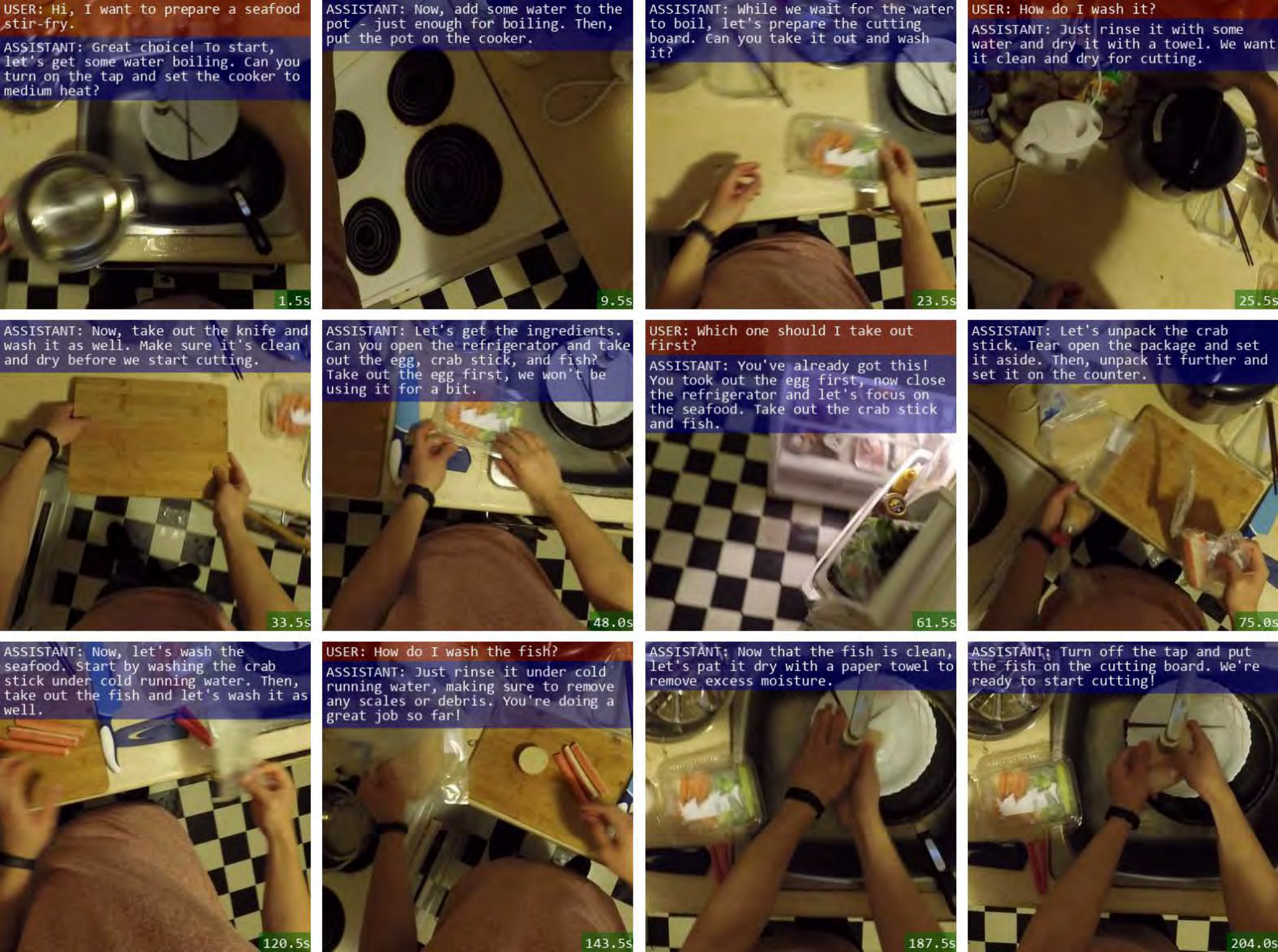}
    \vspace{-8pt}
    \caption{Example of a cooking task in \dataset.\vspace{-8pt}}
    \label{fig:data-sample6}
\end{figure*}

See Figure \ref{fig:data-sample1}-\ref{fig:data-sample6} or example dialogues from \dataset. Due to space constraints, only the first 12 dialogue turns are displayed for each task.

\section{Additional Details of Evaluation Metrics}


\subsection{Pairwise Evaluation}
We utilize the \texttt{all-mpnet-base-v2}\footnote{\url{https://huggingface.co/sentence-transformers/all-mpnet-base-v2}} model from the Sentence-Transformers library \cite{reimers2020sentencebert} to compute text similarity. A similarity threshold of 0.5 is applied to determine correct matches. For temporal cost calculation, the cost-increasing rate $p$ is set to 1.5. The cutoff range $R$ is determined based on the average speaking interval of each dataset: 2.5 seconds for action narration and 1.5–6.0 seconds for dialogue generation. In particular for dialogue generation, we set $L = R/2$ to penalize delayed predictions more heavily, because it is important for the model to provide instruction of the next step before the user begins performing it in the video. 

\vspace{-5pt}
\subsection{LLM-based Evaluation}
\label{supp:llm-eval-detail}
The detailed description of each metric and rubric of 5-scale Likert score is described in List \ref{prompt:llm_eval}, which is also the prompt given to the LLM judge. To reduce the randomness of scoring, we repeat each evaluation for 3 times, and use the average score as the final score for each metric. We use LLaMA-3.1-70B-Instruct as the evaluator in our experiments. 

\begin{lstlisting}[caption={Prompt for LLM-based end-to-end evaluation.},label={prompt:llm_eval},language={}]
You are an expert in evaluating the quality of user-assistant dialogues. Your task is to evaluate dialog responses generated by an assistant model that helps users with their tasks. You should evaluate the dialogs by comparing them to reference gold-standard dialogues from professional assistants.

Requirement:
1. Read dialogues carefully and compare them line by line. Keep you analysis concise and to the point.

2. Evaluate the following aspects: 
- Correctness: does each generated instruction/feedback make sense (correct or relevant) or not, based on the context and the gold-standard reference?
- Promptness: does the assistant provide guidance at the right time, or does it talk too early or too late?
- Efficiency: does the assistant provide the necessary information in a concise and efficient manner, without too much repetition or redundancy information?
- Overall: the overall helpfulness and quality of the assistant's responses.

3. For each aspect, give a score from 1 to 5 based on the following criteria:
- 1=very poor: most of utterances are incorrect, irrelevant, mistimed, inefficient etc
- 2=poor: bad utterances that are incorrect, irrelevant, mistimed are more than good ones
- 3=average: the number of good and bad utterances are roughly the same
- 4=good: more good utterances than bad ones
- 5=excellent: most of utterances are correct, relevant, timely, efficient etc
\end{lstlisting}

\section{Model Implementation Details}


\subsection{VideoLLM-Online Model Implementation}
We use LLaMA-3.1-8B-Instruct \cite{dubey2024llama} as the base LLM and the pretrained SigLIP-SO400M-14-384\footnote{\url{https://huggingface.co/google/siglip-so400m-patch14-384}} model \cite{zhai2023sigmoid} as the frame encoder. To extract frame features, we use the embeddings from the second last layer of the \texttt{[CLS]} token, and $N\times N$ patch features obtained through average pooling of the corresponding patch embeddings. In our experiments, we test with three model variants with $N=0, 1, 2$, resulting in $I=1, 5, 10$ tokens per frame, respectively. We use a two-layer MLP as the projector to project the visual features into the LLM's embedding space following \cite{chen2024videollm}. We remove the separator token between frames because we find it does not help with the performance. 

\begin{table*}[t]
\resizebox{\textwidth}{!}{
\addtolength{\tabcolsep}{-0.1em}
\begin{tabular}{l|c|ccc|ccc|ccc}
\toprule
\multicolumn{1}{c|}{\multirow{2}{*}{Dataset}} & \multicolumn{1}{c|}{\multirow{2}{*}{$\times$}}   & \multicolumn{3}{c|}{I=1}                  & \multicolumn{3}{c|}{I=5}             & \multicolumn{3}{c}{I=10}            \\
\multicolumn{1}{c|}{}                         &   & Ori. Size & Final Size & Proportion      & Ori. Size & Final Size & Proportion & Ori. Size & Final Size & Proportion \\
\midrule
\textbf{Dialogue}                                    &    &           &       & \textbf{47.2\%} &           &      & \textbf{45.9\%}     &           &      & \textbf{46.8\%}     \\
\dataset-Ego4D                                        & 2  & 4795      & 9590       & 6.2\%           & 12350     & 24700      & 9.1\%      & 20718     & 41436      & 9.9\%      \\
\dataset-HoloAssist                                   & 2  & 7645      & 15290      & 9.9\%           & 10957     & 21914      & 8.1\%      & 16116     & 32232      & 7.7\%      \\
\dataset-EgoExoLearn                                  & 2  & 5659      & 11318      & 7.3\%           & 11414     & 22828      & 8.4\%      & 18727     & 37454      & 8.9\%      \\
\dataset-EpicKitchens                                 & 2  & 8051      & 16102      & 10.4\%          & 12901     & 25802      & 9.5\%      & 19320     & 38640      & 9.2\%      \\
\dataset-WTaG                                         & 6  & 929       & 5574       & 3.6\%           & 2051      & 12306      & 4.5\%      & 3537      & 21222      & 5.1\%      \\
\dataset-Assembly101                                  & 2  & 7503      & 15006      & 9.7\%           & 8514      & 17028      & 6.3\%      & 12738     & 25476      & 6.1\%      \\
\hline
\textbf{Summarization}                                &    &           &       & \textbf{14.4\%} &           &       & \textbf{13.2\%}     &           &       & \textbf{13.2\%}     \\
\dataset-Summary                                      & 2  & 11103     & 22206      & 14.4\%          & 17925     & 35850      & 13.2\%     & 27612     & 55224      & 13.2\%     \\
\hline
\textbf{Action Narration}                      &    &           &       & \textbf{18.0\%} &           &       &  \textbf{23.0\%}          &           &      & \textbf{24.2\%}     \\
Ego4D-Narration                              & 1  & 27719     & 27719      & 18.0\%          & 62350     & 62350      & 23.0\%     & 101672    & 101672     & 24.2\%     \\
\hline
\textbf{Auxiliary}                                     &    &           &       & \textbf{20.4\%} &           &       &    \textbf{17.9\%}        &           &       & \textbf{15.8\%}     \\
Something-Something-V2                                     & 10 & 1320      & 13200      & 8.6\%           & 2639      & 26390      & 9.7\%      & 3959      & 39590      & 9.4\%      \\
LLaVA-Pretrain                                        & 2  & 5598      & 11196      & 7.3\%           & 6714      & 13428      & 4.9\%      & 7840      & 15680      & 3.7\%      \\
EgoObjects                                   & 20 & 355       & 7100       & 4.6\%           & 434       & 8680       & 3.2\%      & 552       & 11040      & 2.6\%      \\
\midrule
 \multicolumn{1}{l}{Total}                                        &  \multicolumn{1}{c|}{}   &           & 154k     &                 &           & 271k     &            &           & 420k     &        \\
\bottomrule
\end{tabular}}
\caption{Detailed training data statistics for different model variants under the maximum sequence length of $L=4096$. We report the upsampling ratio ($\times$), the original dataset size after splitting and packing under different $I$, the final dataset size after upsampling, and the proportion of each data source in the final mixture. The final data size grows with the number of tokens used to encode each frame/image.}
\label{tab:training_statistics}
\end{table*}

\subsection{Training}

\paragraph{Training Datasets}
Our models are trained on a mixture of datasets: 
\begin{itemize}
\item \textbf{\dataset}: We use two variants of the dataset, both with and without the task recipe provided as additional knowledge, to enable the model learning and adapting to both setups simultaneously.
\item \textbf{\dataset-Summary}: To enhance summarization capabilities, we construct a video summarization dataset from \dataset. In this dataset, assistant dialogues are removed, leaving only user dialogues and system prompts. The learning objective is to generate the progress summary at the end. This setup requires the model to generate summaries directly from video and user inputs, avoiding reliance on ground-truth assistant dialogue context as a shortcut.
\item \textbf{Online Action Narration}: As described in \S\ref{sec:narration_task}, this task focuses on real-time narration of the camera wearer's actions from live video streams. We reformat the action narration labels from Ego4D \cite{grauman2022ego4d} into the \dataset dialogue style (e.g., "\textit{Assistant: C opens the fridge}"), with "\textit{C}" denoting the camera wearer. To prevent data contamination, we exclude all videos from the validation and test sets of any Ego4D challenges.
\item \textbf{Auxiliary Vision-Language Datasets}: We also incorporate several additional datasets to improve vision-language alignment: image captioning data from LLaVA \cite{liu2024visual}, action recognition from Something-Something-V2\cite{goyal2017something}, and egocentric object detection data from EgoObjects.\cite{zhu2023egoobjects}. We repurposed the labels into dialogue format and train the model to generate them given a specific system prompt for each task.
\end{itemize}

\begin{table*}[h]
\resizebox{0.9\textwidth}{!}{
\addtolength{\tabcolsep}{-0.1em}
\begin{tabular}{lccccccc}
\toprule
               & Ego4D-Narration & Ego4D & HoloAssist & EgoExoLearn & Assembly101 & EpicKitchens & WTaG \\
\midrule
I=1            & 0.3             & 0.3   & 0.3        & 0.3         & 0.3         & 0.2          & 0.3  \\
I=1 (w/ klg)   & 0.3             & 0.3   & 0.3        & 0.3         & 0.3         & 0.2          & 0.4  \\
I=5            & 0.3             & 0.3   & 0.3        & 0.4         & 0.3         & 0.3          & 0.4  \\
I=5 (w/ klg)  & 0.3             & 0.3   & 0.3        & 0.4         & 0.3         & 0.2          & 0.5  \\
I=10           & 0.3             & 0.4   & 0.4        & 0.4         & 0.3         & 0.2          & 0.4  \\
I=10 (w/ klg) & 0.3             & 0.3   & 0.4        & 0.4         & 0.3         & 0.3          & 0.4  \\
\bottomrule
\end{tabular}}
\caption{Selected speaking threshold $\theta$ for each model on each subset. We evaluate a series of $\theta$ values (0.1, 0.2, ...) for each setup and select the optimal threshold based on the appearance of a local maximum in F1 score as $\theta$ increases.}
\label{tab:theta}
\end{table*}

\paragraph{Data Preprocessing}
We extract video frames at a rate of 2 frames per second (FPS) and align the dialogue timestamps with the corresponding frames. The streaming video-dialogue data is preprocessed into sequences of interleaved image and text tokens, as illustrated in Figure \ref{fig:model}. We use a maximum sequence length of $L=4096$ tokens in our experiments. Sequences are constrained to fit within this length, and we aim to make their lengths as close to $L$ as possible to minimize padding and improve computational efficiency. To achieve this, for long videos in \dataset, we split them and inject progress summarization prompts as described in \S\ref{sec:summarization}. For auxiliary vision-language datasets, multiple samples are packed into a single sequence of interleaved image-text format, such as \texttt{<IMAGES><Text><IMAGES><Text>...}. Here, images can consist of single or multiple frames, and text can represent image captions, action descriptions, or object descriptions. This packing strategy significantly reduces the number of samples compared to the original size. Since each frame is encoded into a varying number of tokens (1, 5, or 10), the same number of images can produce different token counts. Consequently, we apply the splitting and packing strategy separately for each setup, where larger $I$ values result in more samples in the final training set. To balance the scale differences among data sources, smaller datasets are upsampled to achieve a more balanced mixture ratio.
The final training data statistics are summarized in Table \ref{tab:training_statistics}. Each data sample comprises a sequence of interleaved image-text tokens, potentially including hundreds of images, which presents significant challenges for data loading during training. To address this, we pre-extract image features using our image encoder and store them on disk. During training, we load these pre-extracted features directly rather than performing feature extraction on-the-fly, resulting in a $6\times$ speedup.

\paragraph{Training Strategy}
We adopt a single-stage training approach following recent practices \cite{karamchetiprismatic,tong2024cambrian,chen2024videollm}. During training, we freeze the image encoder, tune all parameters in the projector layers, and perform parameter-efficient tuning of the LLM using LoRA \cite{hulora} with $r=128$ and $\alpha=256$. The AdamW optimizer \cite{Loshchilov2017DecoupledWD} is used with a learning rate of $2e{-4}$ and 100 warmup steps. We employ a global batch size of 256, 384, and 512 for $I=1$, $5$, and $10$, respectively. All models are trained for 4 epochs on the mixed dataset described above. We use 8$\times$H100 GPUs for training.

\subsection{Inference}
As described in \S\ref{sec:speaking_decision}, a speaking threshold $\theta$ is used to decide whether to speak at each time step, where the model only decides to remain silent if the probability of predicting the \texttt{[EOS]} token exceeds $\theta$. We observe that the quality of model predictions is highly sensitive to the choice of $\theta$. In our experiments, we perform inference multiple times with a series of thresholds on the validation split of each subset to determine the optimal $\theta$ for each model. Given our observation that model performance, in terms of F1 score, follows an inverse U-shaped curve as $\theta$ increases (Figure \ref{fig:pr-tradeoff}), we select the $\theta$ that yields the best local maximum of the F1 score. Table \ref{tab:theta} summarizes the selected $\theta$ values for each subset. While this selection strategy aligns reasonably well with human judgment (as shown in Table \ref{tab:human_pick}), the chosen threshold is optimal for average performance across a set of videos rather than for individual tasks. Additionally, in real-world scenarios, a support set for hyperparameter tuning may not always be available. We leave the development of a better $\theta$ selection strategy for future work.
\section{Human Evaluation}


\paragraph{IRB Approval} The human evaluation process was reviewed and approved by the Institutional Review Board (IRB) of our institution before the experiment started. The participants have all reviewed and signed the consent forms which can be provided upon request.

\paragraph{Synthetic Data Quality Evaluation}
The set of questions and rubrics presented to human evaluators is detailed in List \ref{prompt:human_eval}. The evaluation consists of six questions: four assess the quality of the synthetic dialogues, and two evaluate the accuracy of the generated task goals and recipes. A 4-point Likert scale is used to eliminate a neutral option, encouraging evaluators to express definitive preferences and provide more decisive judgments \cite{garland1991mid}. The evaluation of human-collected dialogues follows the same interface, with evaluators blinded to the source of the dialogue. Each dialogue is independently evaluated by two separate evaluators.

\begin{lstlisting}[caption={Evaluation questions and rubrics for human evaluation on the synthetic data quality of \dataset.},label={prompt:human_eval},language={}]
Q1 (Dialogue Correctness): Are the assistant's instructions or answers factually correct?
1: Incorrect or misleading.
2: Mostly correct but with key errors.
3: Correct with minor issues.
4: Fully correct and precise.

Q2 (Dialogue Helpfulness): How helpful and easy to follow is the assistant's instruction?
1: Confusing and unhelpful.
2: Unnecessary and adds little value.
3: Helpful but hard to follow.
4: Helpful and easy to follow.

Q3 (Dialogue Alignment): Does the dialogue stay aligned with the video content in real-time?
1: Misaligned with the video content.
2: Mostly aligned but with noticeable missteps.
3: Aligned with minor timing issues.
4: Perfectly aligned with the video.

Q4 (Dialogue Naturalness): Does the dialogue sound natural and conversational?
1: Stilted or unnatural.
2: Somewhat natural but awkward in places.
3: Mostly natural with minor awkwardness.
4: Flows naturally and is fully conversational.

Q5 (Task Goal Accuracy): Does the task goal accurately reflect the user's intended task?
1: Inaccurate or completely misses the intended task.
2: Mostly relevant but has key errors.
3: Accurate with minor issues.
4: Fully accurate and precise.

Q6 (Recipe Accuracy): Does the recipe accurately describe the steps needed to accomplish the task?
1: Inaccurate or misleading.
2: Mostly accurate but with notable errors.
3: Accurate with minor errors.
4: Fully accurate and clear.
\end{lstlisting}

\paragraph{Metric Alignment with Human Preference}
For the correlation experiment in Table \ref{tab:human_corr}, we collect generated dialogues from three models: $I=1$, $I=10$, and $I=10$ (w/ klg), on the same set of tasks randomly sampled from our dataset. These generated dialogues are presented side-by-side with the ground-truth dialogues to human evaluators, who are asked to rank the generated dialogues from best to worst, allowing ties. For each task, we derive three pairwise comparison results from the human rankings. Similarly, pairwise comparison results are derived from the rankings obtained using either the F1 score or the LLM-assigned Overall score. Finally, we compute the correlations between these metrics based on the pairwise comparison results. We use a similar evaluation interface for the best-picking experiment in Table \ref{tab:human_pick}, with the only difference being that evaluators are asked to select the single best model instead of ranking all models. The match rate is calculated as the proportion of cases where the best model selected using our proposed metrics aligns with the human selection.

\end{document}